\documentclass[accepted]{uai2022}

\usepackage[round]{natbib}
\usepackage[utf8]{inputenc}
\usepackage[T1]{fontenc}
\usepackage{url,booktabs,nicefrac,microtype}

\usepackage{xcolor}
\definecolor{linkblue}{rgb}{0.1,0.4,0.7}

\usepackage{multicol}
\usepackage{float}
\usepackage{cancel}
\usepackage{physics}
\usepackage{amsfonts,amsmath,amsthm,amssymb}
\usepackage{mathtools}
\usepackage{wrapfig}
\usepackage{pifont}
\usepackage{tikz}
\usetikzlibrary{bayesnet}
\usetikzlibrary{arrows}

\usepackage{wrapfig}

\usepackage{colortbl}
\usepackage{tabularx}
\usepackage{graphbox}
\usepackage{comment}
\usepackage{multirow}
\usepackage{booktabs}
\usepackage{algorithm, algpseudocode, algcompatible}
\usepackage{enumitem} 
\usepackage{caption}
\usepackage{subcaption}
\def\f{\mathbf{f}}

\def\x{\mathbf{x}}
\def\y{\mathbf{y}}
\def\z{\mathbf{z}}
\def\w{\mathbf{w}}
\def\m{\mathbf{m}}

\def\s{\mathbf{s}}
\def\bu{\mathbf{u}}

\def\a{\mathbf{a}}

\def\0{\mathbf{0}}
\def\1{\mathbf{1}}
\def\X{\mathbf{X}}
\def\K{\mathbf{K}}
\def\Z{\mathbf{Z}}
\def\U{\mathbf{U}}
\def\S{\mathbf{S}}
\def\Q{\mathbf{Q}}
\def\Y{\mathbf{Y}}

\def\W{\mathbf{W}}
\def\A{\mathbf{A}}

\newcommand{\bnu}{\boldsymbol\nu}

\newcommand{\bt}{{\boldsymbol{\theta}}}
\newcommand{\bo}{\boldsymbol{\omega}}
\newcommand{\be}{\boldsymbol{\epsilon}}
\newcommand{\bxi}{\boldsymbol{\xi}}

\newcommand{\bphi}{\boldsymbol{\phi}}

\renewcommand{\L}{\mathcal{L}}
\newcommand{\bS}{{\Sigma}}
\newcommand{\bPhi}{\boldsymbol{\Phi}}

\def\R{\mathbb{R}}
\def\E{\mathbb{E}}
\def\N{\mathcal{N}}
\def\GP{\mathcal{GP}}

\def\eye{\textbf{\textrm{I}}}

\DeclareMathOperator{\KL}{KL}

\DeclareMathOperator{\cov}{\mathrm{cov}}

\DeclareMathOperator*{\argmin}{arg\,min}

\title{Variational multiple shooting for Bayesian ODEs with Gaussian processes}

\author[1]{\href{mailto:<pashupati.hegde@aalto.fi>?Subject=Your UAI 2022 paper}{Pashupati~Hegde}{}}
\author[2]{\c{C}a\u{g}atay~Y{\i}ld{\i}z}
\author[1]{Harri~L{\"a}hdesm{\"a}ki}
\author[1]{Samuel~Kaski}
\author[1]{Markus~Heinonen}

\affil[1]{
    Department of Computer Science\\
    Aalto University\\
    Finland
}

\affil[2]{
    University of T\"{u}bingen\\
    Germany
}

\begin{document}
\maketitle
\begin{abstract}
Recent machine learning advances have proposed black-box estimation of \textit{unknown continuous-time system dynamics} directly from data. However, earlier works are based on approximative solutions or point estimates. We propose a novel Bayesian nonparametric model that uses Gaussian processes to infer posteriors of unknown ODE systems directly from data. We derive sparse variational inference with decoupled functional sampling to represent vector field posteriors. We also introduce a probabilistic shooting augmentation to enable efficient inference from arbitrarily long trajectories. The method demonstrates the benefit of computing vector field posteriors, with predictive uncertainty scores outperforming alternative methods on multiple ODE learning tasks.
\end{abstract}

\section{Introduction} \label{section:intro}

Ordinary differential equations (ODEs) are powerful models for continuous-time non-stochastic systems, which are ubiquitous from physical and life sciences to engineering \citep{hirsch2012differential}. In this work, we consider non-linear ODE systems
\begin{align}
    \x(t) &= \x_0 + \int_0^t \f( \x(\tau)) d\tau \label{eq:odeproblemtraj} \\
    \dot{\x}(t) &:= \frac{d\x(t)}{dt} = \f(\x(t)), \label{eq:odeproblem}
\end{align}
where the state vector $\x(t) \in \R^D$ evolves over time $t \in \R_+$ from an initial state $\x_0$ following its time derivative $\dot{\x}(t)$, and $\tau$ is an auxiliary time variable. Our goal is to learn the differential function $\f : \R^D \mapsto \R^D$ from state observations, when the functional form of $\f$ is unknown.

\begin{figure*}[t]
\centering
\includegraphics[width=0.8\textwidth]{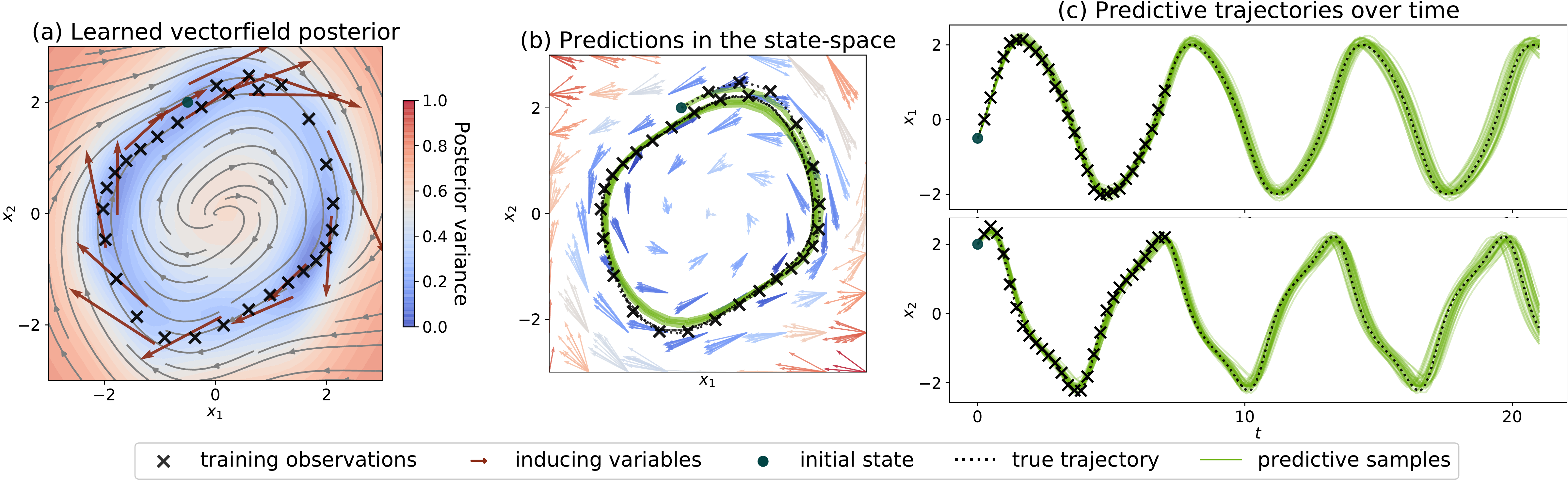}
\caption{Illustration of GPODE: The model learns a GP posterior (a) of a vector field. Valid ODE trajectories are sampled from the posterior process as shown in (b) and (c).}
\label{fig:gpode_illustration}
\end{figure*}

The conventional mechanistic approach involves manually defining the equations of dynamics and optimizing their parameters \citep{butcher2008numerical}, or inferring their posteriors \citep{girolami2008bayesian} from data. However, the equations are unknown or ambiguous for many systems, such as human motion \citep{4359316}. Some early works explored fitting unknown ODEs with splines \citep{henderson2014network}, Gaussian processes \citep{aijo2009learning} or kernel methods \citep{heinonen2014learning} by resorting to less accurate gradient matching approximations \citep{varah1982spline}. Recently, \citet{heinonen2018learning} proposed estimation of free-form non-linear dynamics using Gaussian processes without gradient matching. However, the approach is restricted to learning point estimates of the dynamics, limiting the uncertainty characterization and generalization.  \citet{chen2018neural} proposed modeling ODEs with neural networks and adjoints, which was later extended to the Bayesian setting by \citet{dandekar2020bayesian}. However, the gradient descent training in such approaches can be ill-suited for complex or long-horizon ODEs with typically highly non-linear integration maps \citep{diehl2017}.

In this work, we introduce efficient Bayesian learning of unknown, non-linear ODEs. Our contributions are:
\begin{itemize}
    \item We introduce a way of learning posteriors of vectorfields using Gaussian processes as flexible priors over differentials $\f$, and thereby build on the work by \citet{heinonen2018learning}. We adapt decoupled functional sampling to simulate ODEs from vector field posteriors.
    \item For the difficult problem of gradient optimizations of ODEs, we introduce a novel probabilistic shooting method. It is motivated by the canonical shooting methods from optimal control and makes inference stable and efficient on long trajectories.
    \item We empirically show the effectiveness of the proposed method even while learning from a limited number of observations. We  demonstrate the ability to infer arbitrarily long trajectories efficiently with the shooting extension.
\end{itemize}

\section{Related Works}
\paragraph{Mechanistic ODE models.} In mechanistic modelling the equation $\f_\bt$ is predefined with a set of coefficients $\bt$ to be fitted \citep{butcher2008numerical}. Several works have proposed embedding mechanistic models within Bayesian or Gaussian process models \citep{calderhead2008,dondelinger2013ode,wenk2020odin}. Recently both Julia and Stan have introduced support for Bayesian analysis of parametric ODEs \citep{rackauckas2017differentialequations,stan}. Since this line of work assumes a known dynamics model, we do not consider these methods in the experiments. 

\paragraph{Free-form ODE models.} Multiple works have proposed fitting unknown, non-linear and free-form ODE differentials with gradient matching using splines \citep{ramsay2007parameter}, Gaussian processes \citep{aijo2009learning} or kernel methods \citep{heinonen2014learning}. Recently, \citet{heinonen2018learning} proposed accurate \textit{maximum a posteriori}(MAP) optimisation of vector fields with sensitivity equation gradients \citep{kokotovic1967direct}. Neural ODEs \citep{chen2018neural} introduced adjoint gradients \citep{pontryagin1962mathematical} along with flexible black-box neural network vector fields. Several extensions to learning latent ODEs have been proposed \citep{yildiz2019ode2vae,rubanova2019}. 

\paragraph{Discrete-time state-space models.} There is a large literature on Markovian state-space models that operate over discrete time increments \citep{wang2005,turner2010state, frigola2014variational}. Typically nonlinear state transition functions are modeled with Gaussian processes and applied to latent state estimation or system identification problems with dynamical systems \citep{eleftheriadis2017identification, doerr2018probabilistic, ialongo2019overcoming}. In this paper, we focus strictly on continuous-time models and leave the study of discrete vs. continuous formulations for future work.

\paragraph{Stochastic differential equations.} As an alternative formulation of inferring unknown dynamics from observational data, one can assume stochastic transitions and learn models of stochastic differential equations (SDEs). Existing works have utilized Gaussian processes \citep{archambeau2007gaussian, duncker2019learning, jorgensen2020stochastic} and neural networks \citep{tzen2019neural, li2020scalable} to model non-linear SDEs. However, since they assume a different model (i.e. deterministic transitions vs stochastic transitions), we will restrict the experimental comparisons to other ODE-based approaches.

\section{Methods} \label{section:methods}

We consider the problem of learning ODEs \eqref{eq:odeproblem} with GPs and propose a Bayesian model to infer posteriors over the differential $\f(\cdot)$. 

\subsection{Bayesian modeling of ODEs using GPs}
We assume a sequence of $N$ observations $\Y = (\y_1, \y_2, \ldots \y_N)^T \in \R^{N \cross D}$ along a trajectory, with $\y_i \in \R^D$ representing the noisy observation of the unknown state $\x(t_i) \in \R^D$ at time $t_i$. Similar to \cite{heinonen2018learning}, we assume a zero mean vector-valued Gaussian process prior over $\f$,
\begin{align}
    \f(\x) &\sim \GP(\0, K(\x, \x')),
\end{align}
which defines a distribution of differentials $\f(\x)$ with covariance $\cov[\f(\x), \f(\x')] = K(\x, \x')$, where $K(\x, \x') \in \R^{D \cross D}$ is a stationary matrix-valued kernel. We follow the commonly used sparse inference framework for GPs using inducing variables \citep{titsias2009variational}, and augment the full model with inducing values $\mathbf{U} = (\mathbf{u}_1, \ldots ,\mathbf{u}_M)^T \in \mathbb{R}^{M \times D}$ and inducing locations $\mathbf{Z} = (\mathbf{z}_1, \ldots, \mathbf{z}_M )^T \in \mathbb{R}^{M \times D}$ such that $\mathbf{u}_m = \f(\mathbf{z}_m)$. The inducing variables are trainable `landmark' state-differential pairs, from which the rest of the differential field is interpolated (See Figure \ref{fig:gpode_illustration}, where arrow locations are the $\mathbf{z}_m$ and arrow end-points are the $\mathbf{u}_m$). The inducing augmentation leads to the following prior and conditionals \citep{hensman2013gaussian}:
\begin{align}
    p(\U) &= \N(\U | \0, \K_{\Z\Z}), \\
    p(\f | \U; \Z) &= \N(\f | \A \mathrm{vec}(\U), \K_{\X \X} - \A \K_{\Z \Z} \A^T),
\end{align}
where $\X = (\x_1, \x_2, \ldots \x_{N'})^T \in \R^{N' \times D}$ collects all the intermediate state evaluations $\x(t_i)$ encountered along a numerical approximation of the true continuous ODE integral \eqref{eq:odeproblemtraj}, $\f = (\f(\x_1)^T, \ldots, \f(\x_{N'})^T)^T \in \mathbb{R}^{N'D \times 1}$, $\K_{\X\X}$ is a block-partitioned matrix of size $N'D \times N'D$ with $D \times D$ blocks, so that block $(\K_{\X\X})_{i,j} = K(\x_i, \x_j)$, and $\A = \K_{\X\Z} \K^{-1}_{\Z\Z}$. For notational simplicity, we assume that  the measurement time points are among the time points of the intermediate state evaluations of a numerical ODE solver.

The joint probability distribution follows
\begin{align}
    p(\Y, \f, \U, \x_0) &= p(\Y|\f,\x_0)p(\f,\U)p(\x_0) \\
    &\hspace{-10mm} = \prod\limits_{i=1}^{N}p(\y_i|\f, \x_0) p(\f|\U) p(\U) p(\x_0),
\end{align}
where the conditional distribution $p(\y_i| \f, \x_0) = p(\y_i|\x_i)$ computes the likelihood over ODE state solutions $\x_i = \x_0 + \int_0^{t_i} \f( \x(\tau)) d\tau$.

\subsection{Variational inference for GP-ODEs}
In contrast to earlier approach that estimates MAP solutions \citep{heinonen2018learning}, our goal is to infer the posterior distribution $p(\f,\x_0 | \Y)$ of the vector field $\f$ and initial value $\x_0$ from observations $\Y$. The posterior is intractable due to the non-linear integration map $\x_0 \overset{\f}{\mapsto} \x(t)$. 

We use the stochastic variational inference (SVI) formulation for sparse GPs \citep{hensman2013gaussian} in this work. We introduce a factorized Gaussian posterior approximation  for the inducing variables across state dimensions $q(\U) = \prod_{d=1}^{D}\N(\bu_d|\m_d, \Q_d), \bu_d \in \R^M$ where $\m_d \in \R^M, \Q_d \in \R^{M\times M}$ are the mean and covariance parameters of the variational Gaussian posterior approximation for the inducing variables. We treat the inducing locations $\Z$ as optimized hyperparameters. The posterior distribution for the variational approximation can be written as
\begin{align}
    q(\f) &= \int p(\f|\U) q(\U) d\U \\
     &\hspace{-8mm}= \int \N\left(\f | \A \mathrm{vec}(\U), \K_{\X\X} - \A \K_{\Z\Z}\A^T \right) q(\U) d\U. \label{eq:inducing_posterior_gp}
     \raisetag{2\normalbaselineskip}
\end{align}

The posterior inference goal then translates to estimating the posterior $p(\f, \U, \x_0 | \Y)$ of the inducing points $\U$ and initial state $\x_0$. Under variational inference this learning objective
\begin{align}
    \argmin_{q} \: \KL\big[ \, q(\f,\U,\x_0) \, || \, p(\f,\U,\x_0|\Y) \, \big]
\end{align}
translates into maximizing the evidence lowerbound (ELBO), 
\begin{align}
    \log p(\Y) &\ge \sum_{i=1}^N \overbrace{\E_{q(\f, \x_0)} \log p(\y_i | \f, \x_0)}^{\text{variational likelihood}}  - \overbrace{\KL[ q(\U) || p(\U)]}^\text{inducing KL} \notag \\
    &\quad - \underbrace{\KL[ q(\x_0) || p(\x_0)]}_\text{initial state KL},
\end{align}
where we also assume variational approximation $q(\x_0) = \N(\a_0, \bS_0)$ for the initial state $\x_0$. See supplementary section 1.1 for detailed derivations of the above equations.

\begin{figure*}[!h]
    \centering
    \begin{subfigure}[b]{0.9\columnwidth}
    \centering
    \includegraphics[width=1.0\textwidth]{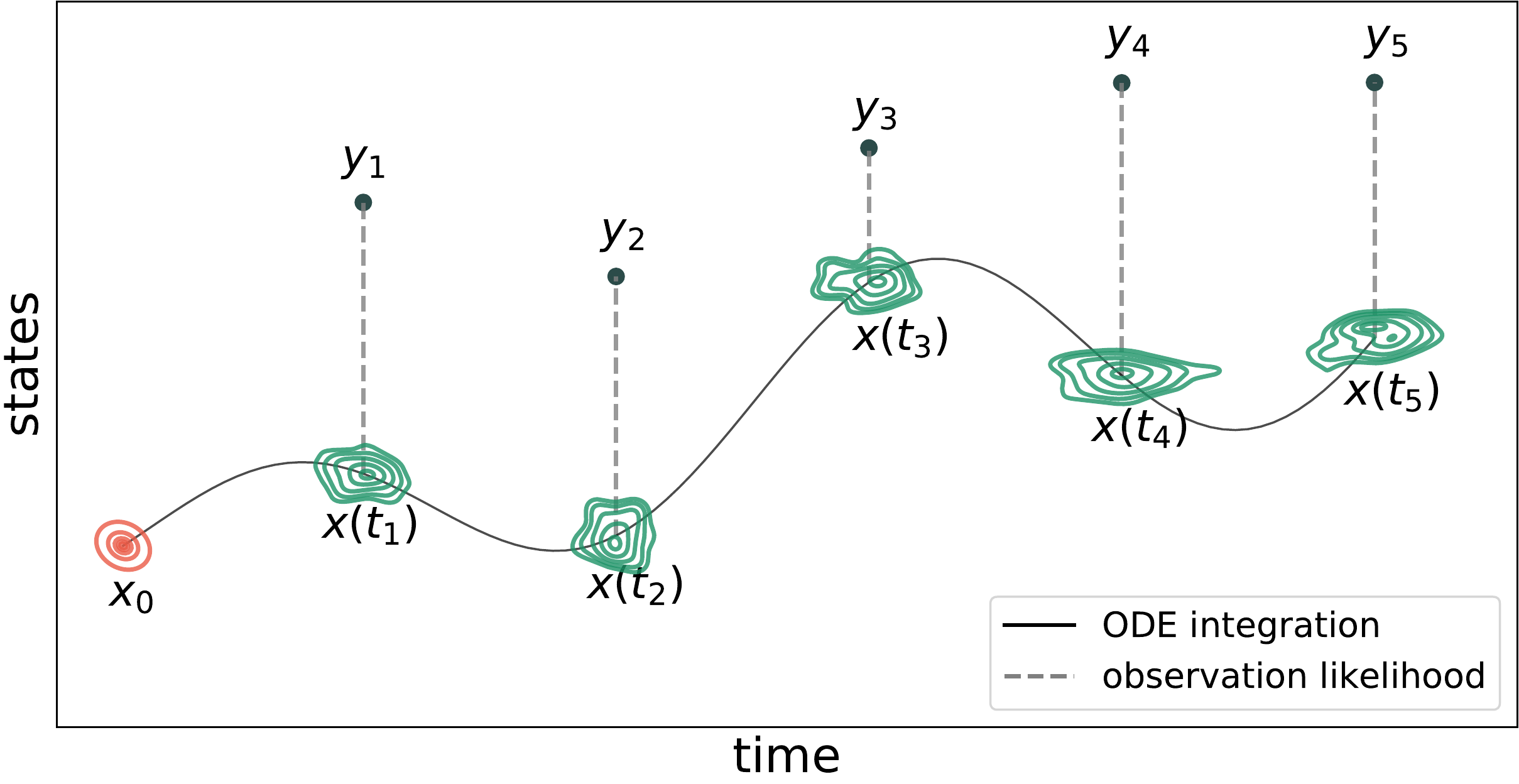}
    \caption{The full model formulation}
    \end{subfigure}
    \qquad
    \begin{subfigure}[b]{0.9\columnwidth}
    \centering
    \includegraphics[width=1.0\textwidth]{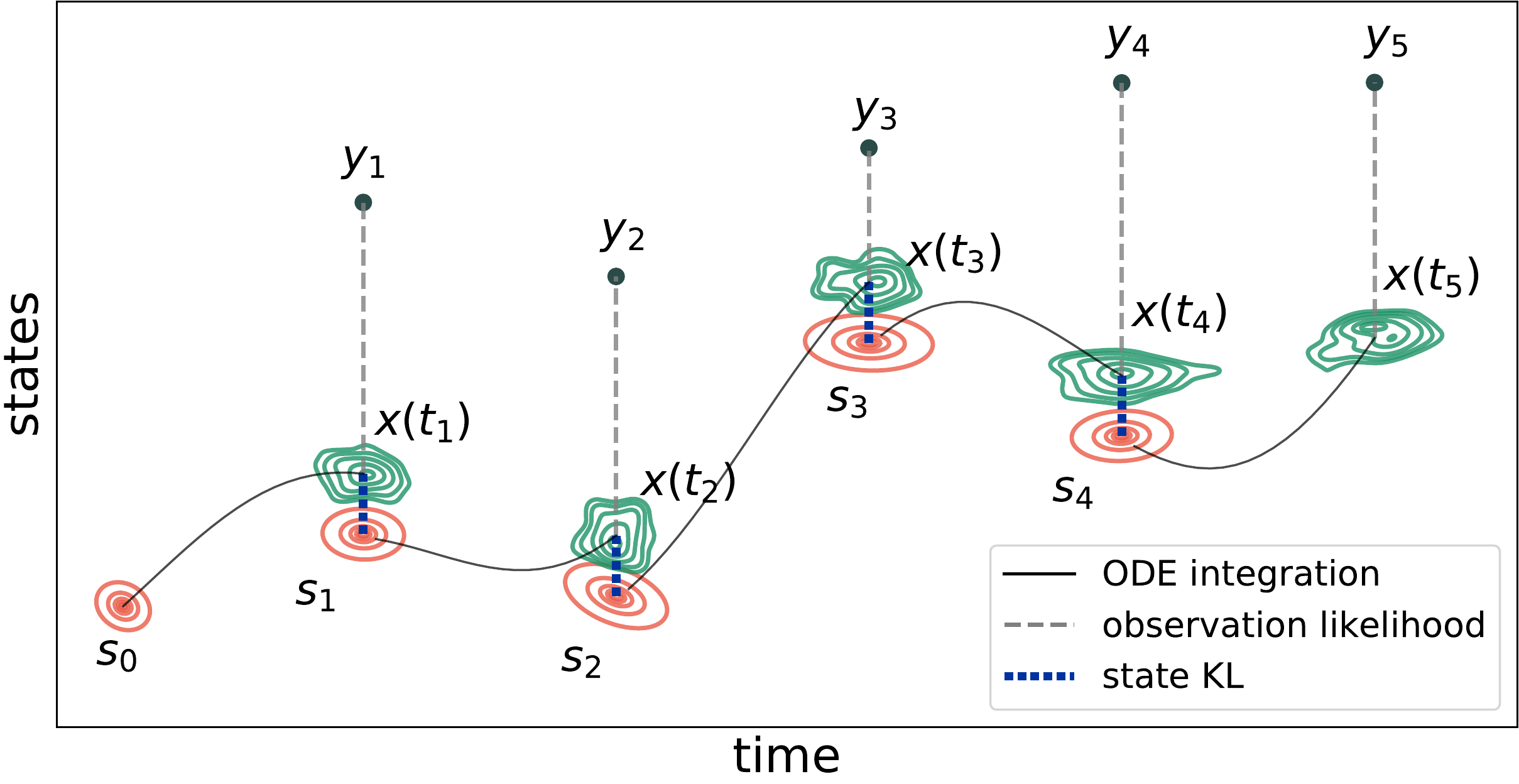}
    \caption{Shooting augmentation}
    \end{subfigure}
    \caption{Illustrations of GPODE formulations: the full model formulation (a) follows the long trajectory integration, whereas the shooting version (b) splits the long trajectory into multiple short subintervals.}
    \label{fig:shooting_illustration}
\end{figure*}

\subsection{Sampling ODEs from Gaussian processes}

The Picard-Lindel\"{o}f theorem \citep{lindelof} ensures valid ODE systems define unique solutions to the initial value problem (IVP) \eqref{eq:odeproblemtraj}. In order to sample valid state trajectories for the IVP, we need to efficiently sample GP functions $\f(\cdot) \sim q(\f)$ \eqref{eq:inducing_posterior_gp}. This way, we can evaluate the sample function $\f(\x(t))$ at arbitrary states $\x(t)$ encountered during ODE forward integration, while accounting for both the inducing and interpolation distributions of Equation \eqref{eq:inducing_posterior_gp}.  Unfortunately, function-space sampling of such GPs has prohibitive cubic complexity \citep{rasmussen2006gaussian,pmlr-v108-ustyuzhaninov20a}, while the more efficient weight-space sampling with Fouriers cannot accurately express the posterior \eqref{eq:inducing_posterior_gp} \citep{wilson2020efficiently}. 

We use the decoupled sampling that decomposes the posterior into two parts \citep{wilson2020efficiently},
\begin{align}
\label{eq:decoupled_conditional}
    \overbrace{\f(\x)|\U}^\text{posterior} &= \overbrace{\f(\x)}^\text{prior} + \overbrace{K(\x,\Z)K(\Z,\Z)^{-1}(\U - \f_{\Z}))}^\text{update}.  \\
    &\approx \sum_{i=1}^{F} \w_i \bphi_i(\x) + \sum_{j=1}^{M} \bnu_j K(\x, \z_j),
\end{align}
where we use $F$ Fourier bases $\bphi_i(\cdot)$ with $\w_i \sim \N(\0,I)$ \citep{rahimi2007random} to represent the stationary prior, and function basis $K(\cdot,\z_j)$ for the posterior update with $\bnu = K(\Z,\Z)^{-1}(\U - \bPhi \W)$, $\bPhi = \bphi(\Z) \in \R^{M \times F}, \W \in \R^{F \times D}$. By combining these two steps, we can accurately evaluate functions from the posterior \eqref{eq:inducing_posterior_gp} in linear time at arbitrary locations. We refer the reader to the supplementary section 1.2 for more details.  We note that concurrent works by \citet{mikheeva2021aligned} and \citet{ensinger2021symplectic} also utilize the decoupled-sampling to infer ODE posteriors with GPs. 

\begin{figure*}[t]
    \centering
    \includegraphics[width=0.8\textwidth]{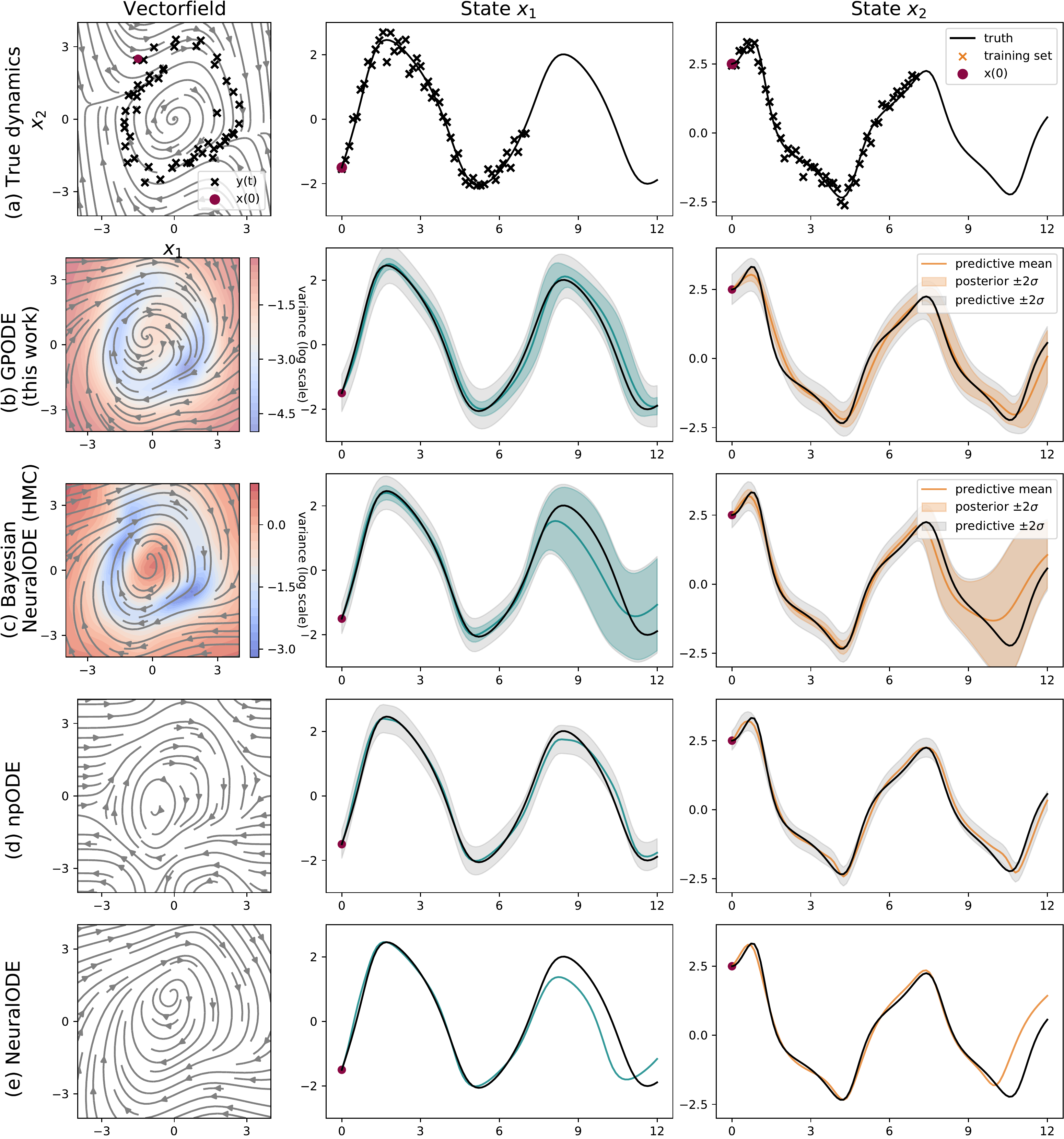}
    \caption{Learning the 2D Van der Pol dynamics \textbf{(a)} with alternative methods \textbf{(b-d)}. Column 1 shows the vector fields while columns 2 and 3 show the state trajectories $x_1(t)$ and $x_2(t)$. GPODE learns the posterior accurately.}
    \label{fig:vdp_illustration}
\end{figure*}

\subsection{Augmenting the ODE model with shooting system}
A key bottleneck in ODE modeling is the poor gradient descent performance over long integration times $\x_{0:T}$, which can exhibit vanishing or exploding gradients \citep{haber2017,choromanski2020}. Earlier approaches tackled this issue mainly with more accurate numerical solvers \citep{zhuang2020,zhuang2021mali}. The nonlinearity of the integration map $\x_0 \overset{\f}{\mapsto} \x_t$ motivates us to instead segment the full integration $\x_{0:T}$ into short segments, which are easier to optimize and can be trivially parallelized. This is called the \emph{multiple shooting} method in optimal control literature \citep{osborne1969,bock1984}, in the context of parameter estimation of ODEs \citep{vandomselaar1975nonlinear,bock1983recent}. Recently, \citet{massaroli2021differentiable, turan2021multiple} also introduced a multiple-shooting framework within the context of deterministic neural ODEs. We introduce probabilistic shooting for the Gaussian process posterior inference of ODEs.

We begin by introducing shooting state variables $\mathbf{S} = (\s_0,\s_1,\ldots,\s_{N-1})$, $\s_i \in \R^D$, and segment the continuous state function $\x(t;\x_0)$ \eqref{eq:odeproblemtraj} into $N$ segments $\{(\s_{i-1},\x(t_i; \s_{i-1}))\}_{i=1}^{N}$ that branch from the shooting variables $\s_{i-1}$ (See Figure \ref{fig:shooting_illustration});
\begin{align}
    \x(t_i; \s_{i-1}) &= \s_{i-1} + \int_{t_{i-1}}^{t_i} \f(\x(\tau)) d\tau  \label{eq:shooting_ivps}.
\end{align}
In addition, every shooting variable is approximately matched with the ODE state evolution from the previous shooting state,
\begin{align}
    \s_i &= \x(t_i; \s_{i-1}) + \bxi \label{eq:shooting_constraints},
\end{align}
where $\bxi \in \R^D$ represents the tolerance parameter controlling the shooting approximation. The augmented system is equivalent to the original ODE system in case the constraints $\s_i = \x(t_i; \s_{i-1})$ are satisfied exactly at the limit $\bxi \to \0$. We place a Gaussian prior over the tolerance parameter $\bxi \sim \N(\0, \sigma^2_\xi\eye)$, which translates into the following prior over shooting variables  
\begin{align}
    p(\s_i|\s_{i-1}) &= \N(\s_i|\x(t_i; \s_{i-1}), \sigma^2_\xi \eye). \label{eq:shooting_prior}
\end{align}

Further, the joint probability of the augmented model after placing a GP prior over the vectorfield $\f$ can be written as 
\begin{align}
p(\Y,\S,\f) &= \prod\limits_{i=1}^{N} p(\y_{i}|\s_{i-1}, \f) \prod\limits_{i=1}^{N-1}p(\s_{i}|\s_{i-1}, \f)p(\s_0) p(\f).
\end{align}

\subsection{Variational inference for the augmented model}
To infer the augmented posterior $p(\f,\U, \S| \Y)$ we introduce variational approximation for the shooting variables $q(\S) = q(\s_0) \cdots q(\s_{N-1})$, where each distribution $q(\s_i) = \N(\s_i|\a_i,\bS_i)$ is a Gaussian. This results in the joint variational approximation
\begin{align}
    q(\S,\f,\U) &= \prod_{i=0}^{N-1} q(\s_i) p(\f|\U) q(\U),
\end{align}
and the following evidence lower bound for the shooting model,
\begin{align}
    \L_{\mathrm{shooting}}
    &= \sum_{i=1}^N \E_{q(\s_{i-1},\f)} \Big[\log p(\y_i|\s_{i-1}, \f)\Big] \nonumber\\
    &\hspace{-15mm} + \sum_{i=1}^{N-1} \E_{q(\s_{i},\s_{i-1},\f)} \Big[ \log p\left(\s_i | \s_{i-1}, \f \right) \Big] - \E_{q(\s_i)}  \Big[ \log q(\s_{i})  \Big] \nonumber\\
    &\hspace{-15mm}  - \KL[q(\s_0) \, || \, p(\s_0)] - \KL[ q(\U) \, || \, p(\U)]. \label{eq:shooting_elbo}
\end{align}

 The ELBO consists of an expected log-likelihood term, which matches the state evolution \eqref{eq:shooting_ivps} from every shooting variable to the corresponding observation. In addition, the posterior approximation for every shooting variable is also matched with the ODE evolution of the approximated posterior of the previous shooting state, leading to corresponding cross-entropy and entropy terms. 
 
The ELBO for the augmented shooting model requires solving only the short segments $\eqref{eq:shooting_ivps}$ with simpler integration maps, thus great at mitigating problems with vanishing/exploring gradients. Since the involved numerical ODE integration can be done in parallel, the shooting model is also computationally faster than the full model in practice. See supplementary section 1.3 for a plate diagram and detailed derivation of the approach.

\begin{table*}[t]
\centering
\caption{VDP system learning performance on extrapolation task with observations on regular (task 1) and irregular time intervals (task 2). We report mean $\pm$ standard error over 5 runs from different random initialization, the best values bolded. ($\uparrow$): higher is better, ($\downarrow$) lower is better}
\resizebox{0.8\textwidth}{!}{
\begin{tabular}{l c c c c}
\toprule
 & \multicolumn{2}{c}{Task 1: Regular time-grid} & \multicolumn{2}{c}{Task 2: Irregular time-grid}\\
 \cmidrule(lr){2-3} 
 \cmidrule(lr){4-5} 
 & MNLL ($\downarrow$)& MSE ($\downarrow$)& MNLL ($\downarrow$) & MSE ($\downarrow$) \\
\midrule
Bayesian NeuralODE (HMC) &  $0.82 \pm 0.01$ & $1.45 \pm 0.04$ & $0.88 \pm 0.01$ & $1.68 \pm 0.04$ \\
NeuralODE & - & $0.29 \pm 0.11$ & - & $0.55 \pm 0.07$\\
npODE & $1.47 \pm 0.59 $ & $0.16 \pm 0.05 $ & $8.89 \pm 3.06 $ & $2.08 \pm 0.78 $ \\
GPODE & $\mathbf{0.60 \pm 0.03}$ & $\mathbf{0.13 \pm 0.01}$ & $\mathbf{0.41 \pm 0.18}$ & $\mathbf{0.21 \pm 0.07}$ \\
\bottomrule
\end{tabular}
}
\label{table:vdp_illustration}
\end{table*}

\section{Experiments} \label{section:experiments}
We validate the proposed method on Van der Pol (VDP) and FitzHugh–Nagumo (FHN) systems and on the task of learning human motion dynamics (MoCap). The predictive performance of the proposed GPODE is compared against npODE \citep{heinonen2018learning}, NeuralODE \citep{chen2018neural} and Bayesian version of NeuralODE \citep{dandekar2020bayesian}. We use 16 inducing points in VDP and FHN experiments and 100 inducing points for the MoCap experiments. Except for the NeuralODE model, we assume Gaussian observation likelihood, and infer the unknown noise scale parameter from the training data. All the experiments use squared exponential kernel with automatic relevance determination (ARD) along with 256 Fourier basis functions for decoupled GP sampling. Along with the variational parameters, kernel lengthscales, signal variance, noise scale, and inducing locations are jointly optimized against the model ELBO while training. In addition, for the shooting model, we fix the constraint tolerance parameter to a small value $\sigma^2_\xi = 1e^{-6}$ consistently across all the experiments. In all the shooting experiments, we considered the number of shooting segments to be the same as the number of observation segments in the dataset. A codebase for implementing the proposed methods is provided \url{https://github.com/hegdepashupati/gaussian-process-odes}. 

We use the \texttt{dopri5} solver with tolerance parameters \texttt{rtol}$=1e^{-5}$ and \texttt{atol}$=1e^{-5}$, and use the adjoint method for computing loss gradients with \texttt{torchdiffeq}\footnote{\url{https://github.com/rtqichen/torchdiffeq}} package \citep{chen2018neural}. All the experiments are repeated 5 times with random initialization, and means and standard errors are reported over multiple runs. The predictive performance of different models are measured with mean squared error (MSE) and mean negative log likelihood (MNLL) metrics. 

\subsection{Learning Van der Pol dynamics}
We first illustrate the effectiveness of the proposed method by inferring the vector field posterior on a two-dimensional VDP (see Figure \ref{fig:vdp_illustration}),
\begin{align}
    \dot{x}_1 = x_2, \\ \nonumber
    \dot{x}_2 = -x_1 + 0.5 x_2 (1-x_1^2).
\end{align}
We simulate a trajectory of 50 states following the true system dynamics from the initial state $\left(x_1(0), x_2(0)\right) = \left(-1.5, 2.5 \right)$, and add Gaussian noise with $\sigma^2=0.05$ to generate the training data. We explore two scenarios with training time interval $t \in [0,7]$ and forecasting interval $t \in [7,14]$: (1) over a regularly sampled time grid, (2) over an irregular grid using uniform random sampling of time points. Task (2) demonstrates one of the key advantages of continuous-time models with the ability to handle irregular data. 

Figure \ref{fig:vdp_illustration}(b) shows that both GPODE and Bayesian NeuralODE learn a vector field posterior whose posterior mean closely matches the ground truth, with low variance (\textcolor{blue}{blue regions}) near the observed data. The posterior variance increases away from the observed data (\textcolor{orange}{orange regions}), indicating a good uncertainty characterization, while the npODE with MAP estimation seems to overfit. NeuralODE learns an appropriate vector field, but requires careful tuning of regularization and hyperparameters for a good fit with a limited number of observations. A quantitative evaluation of the model fits in Table \ref{table:vdp_illustration} indicates the better performance of GPODE as compared to the other methods under comparison.

\begin{table}
\caption{Imputation results on the FHN system.}
\centering
\resizebox{0.9\columnwidth}{!}{
\begin{tabular}{lcc}
\toprule
 & MNLL ($\downarrow$) & MSE ($\downarrow$)  \\
\midrule
Bayesian NeuralODE (HMC) & $0.77 \pm 0.12$ & $0.24 \pm 0.03$ \\
NeuralODE & - & $0.18 \pm 0.00$ \\
npODE & $6.49 \pm 1.49$ & $\mathbf{0.08 \pm 0.01}$ \\
GPODE & $\mathbf{0.09 \pm 0.05}$ & $\mathbf{0.07 \pm 0.02}$ \\
\bottomrule
\end{tabular}
}
\label{table:fhn_interpolation}
\end{table}

\subsection{Learning with missing observations}
We illustrate the usefulness of learning Bayesian ODE posteriors under missing data with the FHN oscillator
\begin{align}
\dot{x}_1 &= 3(x_1 - x_1^3/3 + x_2), \\ \nonumber
\dot{x}_2 &= (0.2 - 3x_1 - 0.2 x_2)/3.   
\end{align}
We generate a training sequence by simulating 25 regularly-sampled time points from $t \in [0, 5.0]$ with added Gaussian noise with $\sigma^2 = 0.025$. We remove all observations at the quadrant $x_1>0, x_2<0$ and evaluate model accuracy in this region. The interpolation performance for different models is shown in Table \ref{table:fhn_interpolation}. The point estimates of npODE and NeuralODE have biases, while the Bayesian variants
of GPODE and NeuralODE provide good uncertainty estimates corresponding to their better predictive performance. 

\begin{figure}
\centering
\includegraphics[width=0.85\columnwidth]{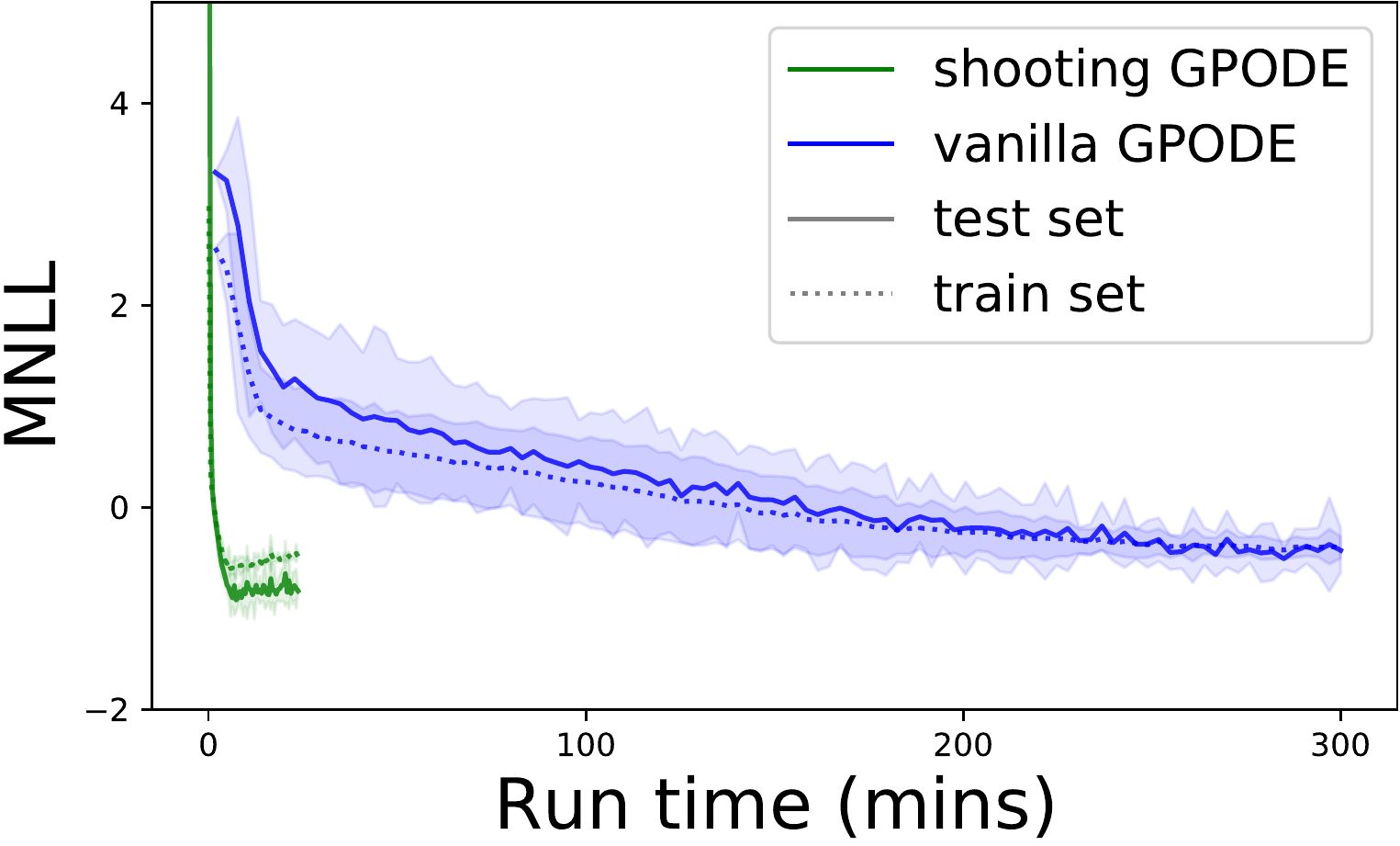}
\caption{Optimization efficiency with GPODE models.}
\label{fig:gpode_shooting_efficiency}
\end{figure}

\begin{figure*}[t]
    \centering
    \includegraphics[width=0.9\textwidth]{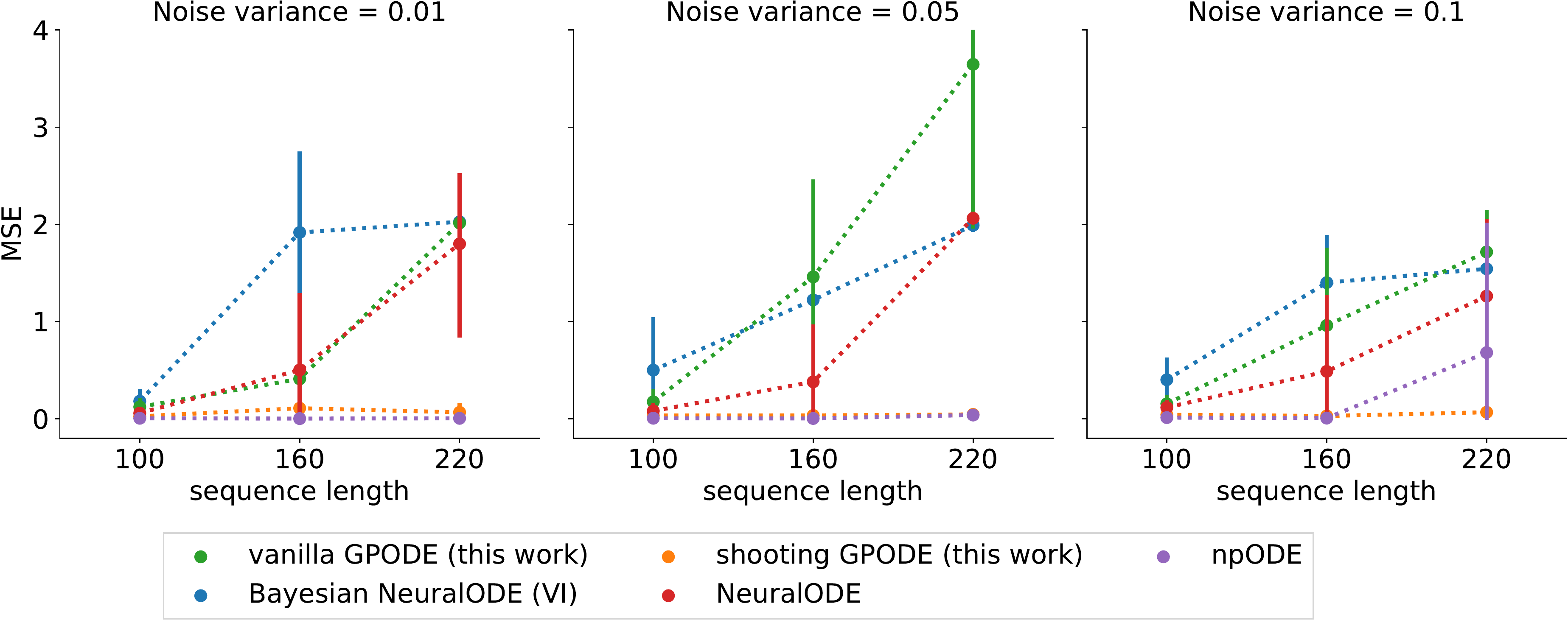}
    \caption{Varying sequence length and observation noise: shooting formulation makes GPODE feasible for long sequences, outperforming the non-shooting version and competing methods. We report the results for different levels of observation noise and training sequence length on the VDP system.}    
    \label{fig:seqlen_illustration}
\end{figure*}

\subsection{Learning long trajectories with the shooting formulation}
We demonstrate the necessity of the shooting formulation for working with long training trajectories. We use the VDP system with four observations per unit of time for $T = (25,40,55)$ corresponding to $N = (100,160,220)$ observed states. We also vary the observation variance as $\sigma^2 = (0.01,0.05,0.1)$ and test the model for forecasting additional 50 time points.

Figure \ref{fig:seqlen_illustration} demonstrates that vanilla-GPODE and NeuralODE, and Bayesian NeuralODE fail to fit the data with long sequences on all noise levels. In contrast, inference for the shooting model is successful in all settings. The npODE is remarkably robust to long trajectories. We believe the robustness of npODE mainly stems from the excellent parameter initialization strategy (see supplementary section 2.2) coupled with the fully deterministic optimization setup (no reparametrization gradients).

Figure \ref{fig:gpode_shooting_efficiency} shows a runtime trace comparison between vanilla GPODE and the shooting variant in wall-clock time for a fixed budget of 15000 optimization steps on the VDP system with $N=100$, $T=25$ and $\sigma^2=0.01$. The shooting model converges approximately 10 times faster. The speedup stems from the parallelization of the shooting ODE solver, since the shooting method splits the full IVP problem into numerous short and less non-linear IVPs. In addition, the shooting method relaxes the inference problem with its auxiliary augmentation.  This experiment was conducted on a system with AMD Ryzen 5 3600 processor and Nvidia GeForce GTX 1660S GPUs.

\begin{figure*}[t]
    \centering
    \includegraphics[width=0.85\textwidth]{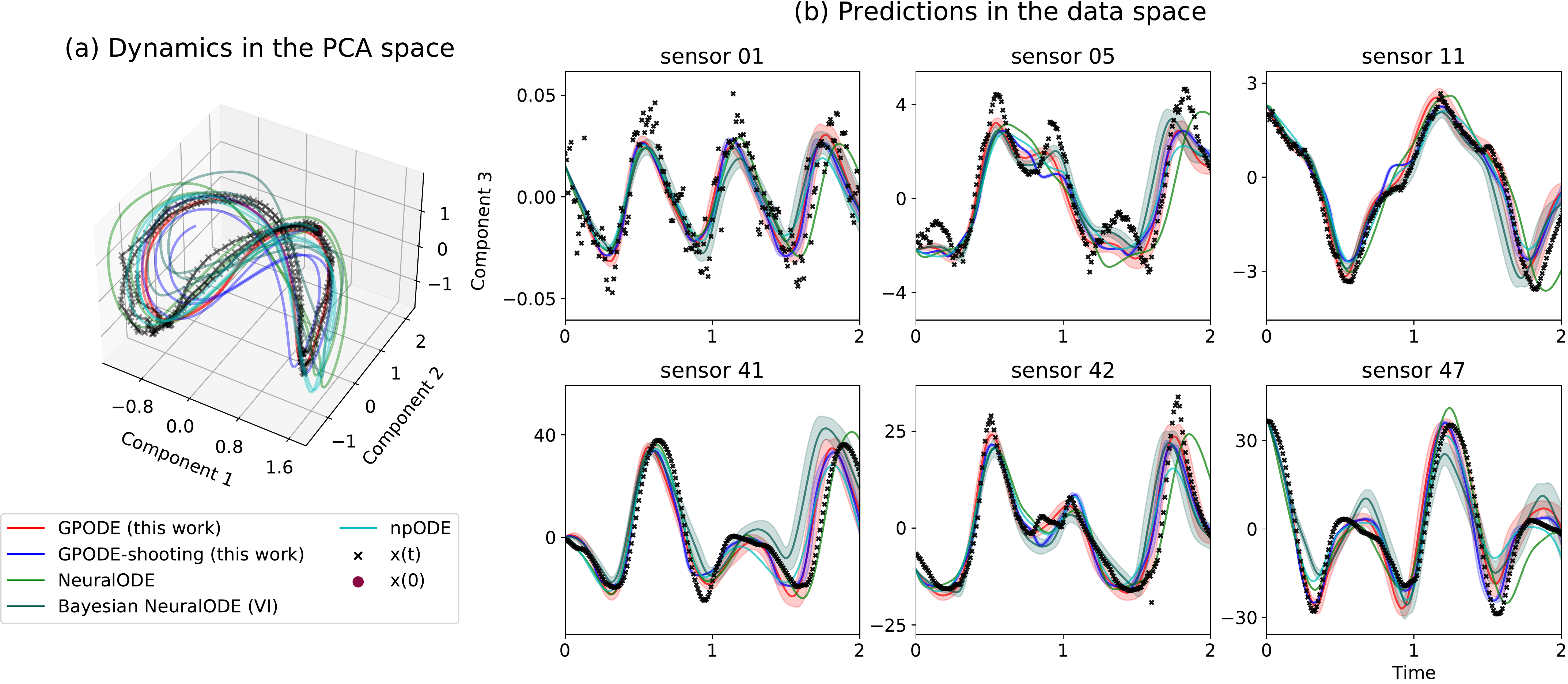}
    \caption{Learning the walking dynamics of subject \texttt{39}: The true dynamics and predicted dynamics (mean) for the first three components in PCA space are shown in (a). Corresponding trajectories in the observation space for 6 different sensors are shown in (b) (We do not plot the observation noise variance)}
    \label{fig:mocap_illustration}
\end{figure*}

\begin{table*}
\centering
\caption{Test MNLL and MSE metrics for dynamics prediction task on CMU MoCap dataset.}
\resizebox{0.9\textwidth}{!}{
\begin{tabular}{c l  c c  c c  c c}
\toprule
\multirow{2}*{Metric}  & \multirow{2}*{Method} & \multicolumn{2}{c}{Subject 09}
 & \multicolumn{2}{c}{Subject 35}
 & \multicolumn{2}{c}{Subject 39} \\
 \cmidrule(lr){3-4}
 \cmidrule(lr){5-6}
 \cmidrule(lr){7-8}
 & & short & long
 & short & long
 & short & long \\
\midrule
\multirow{3}*{MNLL($\downarrow$)}
& Bayesian NeuralODE (VI) & $2.03 \pm 0.10 $ & $1.50 \pm 0.05 $ & $1.42 \pm 0.05 $ & $1.37 \pm 0.06 $ & $1.61 \pm 0.07 $ & $1.45 \pm 0.03 $ \\
& npODE & $2.09 \pm 0.01 $ & $1.78 \pm 0.08 $ & $1.67 \pm 0.02 $ & $1.66 \pm 0.04 $ & $2.06 \pm 0.05 $ & $1.78 \pm 0.04 $ \\
& GPODE-vanilla & $1.30 \pm 0.02$ & $1.26 \pm 0.02$ & $1.27 \pm 0.04$ & $1.39 \pm 0.04$ & $1.29 \pm 0.01$ & $\mathbf{1.13 \pm 0.01}$ \\
& GPODE-shooting & $\mathbf{1.19 \pm 0.02}$ & $\mathbf{1.14 \pm 0.02}$ & $\mathbf{1.25 \pm 0.06}$  &  $\mathbf{1.08 \pm 0.02}$ & $\mathbf{1.25 \pm 0.01}$ & $1.36 \pm 0.02$  \\

\midrule
\multirow{3}*{MSE($\downarrow$)} 
& Bayesian NeuralODE (VI) & $25.50 \pm 1.70 $ & $21.32 \pm 2.58 $ & $23.09 \pm 3.95 $ & $20.86 \pm 2.95 $ & $53.34 \pm 5.31 $ & $39.66 \pm 6.82 $ \\
& NeuralODE & $27.53 \pm 2.87 $ & $33.83 \pm 2.46 $ & $36.50 \pm 3.86 $ & $23.54 \pm 0.56 $ & $115.38 \pm 10.96 $ & $53.51 \pm 2.98 $ \\
& npODE & $17.91 \pm 1.62 $ & $19.76 \pm 4.29 $ & $26.24 \pm 2.88 $ & $22.83 \pm 3.91 $ & $92.80 \pm 15.74 $ & $55.94 \pm 4.63 $ \\
& GPODE-vanilla & $15.78 \pm 0.67$ &  $12.62 \pm 1.14$  & $16.14 \pm 0.99$ & $15.53 \pm 0.76$ & $\mathbf{20.71 \pm 1.25}$ & $23.64 \pm 1.94$ \\
& GPODE-shooting  & $\mathbf{9.11 \pm 0.37}$  & $\mathbf{8.38 \pm 1.23}$ & $\mathbf{10.11 \pm 0.79}$ & $\mathbf{11.66 \pm 0.73}$ & $26.72 \pm 0.63$ & $\mathbf{21.17 \pm 2.88}$  \\
\bottomrule
\end{tabular}
}
\label{table:mocap}
\end{table*}

\subsection{Learning human motion dynamics}
We learn the dynamics of human motion from noisy experimental data from CMU MoCap database for three subjects, \texttt{09}, \texttt{35} and \texttt{39}. The dataset consists of 50 sensor readings from different parts of the body while walking or running. We follow the preprocessing of \citet{4359316} and center the data. The dataset was further split into train, test, and validation sequences. We observed that the NeuralODE, the Bayesian NeuralODE version with VI, and npODE models suffer from over-fitting, and we remedy this by applying early stopping by monitoring the validation loss during optimization.

We project the original 50-dimensional data into a 5-dimensional latent space using PCA and learn the dynamics in the latent space. To compute the data likelihood, we project the latent dynamics back to the original data space by inverting the PCA. We divide the experiment into sub-tasks MoCap-short and MoCap-long, based on the length of the sequence considered for model training (see the supplementary section for more details on the dataset and experimental setup). The model predictive performance is measured on unseen test sequences in both tasks. 

Table \ref{table:mocap} indicates that GPODE outperforms the competing npODE and NeuralODE model variants. Figure \ref{fig:mocap_illustration} visualizes the predicted dynamics for a test sequence. The GPODE variants have reasonable posterior uncertainties, while NeuralODE variants and npODE tend to be overconfident and make more mistakes (see Figure \ref{fig:mocap_illustration} (b), sensors \texttt{05}, \texttt{41} and \texttt{47}) . We note that some variations in the data space cannot be accurately estimated due to the low-dimensional PCA projection. 

\section{Conclusion and Discussion}
We proposed a novel model for Bayesian inference of ODEs using Gaussian processes. With this approach, one can model unknown ODE systems directly from the observational data and learn posteriors of the continuous-time vector fields. In contrast, earlier works produce point estimate solutions. We believe this to be a significant addition to the data-descriptive ODE modeling methods, especially for applications where uncertainty quantification is critical. Many conventional machine learning algorithms have been interpreted and modeled as continuous-time dynamical systems, with applications to generative modeling \citep{grathwohl2019ffjord} and probabilistic alignment \citep{pmlr-v108-ustyuzhaninov20a}, among others. However, scaling GPs to high-dimensional datasets (such as images) can be a bottleneck. The applicability of the proposed model as a plug-in extension for these applications can be studied as part of future work. 

We also highlighted a problem of learning black-box ODE models on long trajectories and proposed a probabilistic shooting framework enabling efficient inference on such tasks. This framework can be applied to other existing approaches, such as NeuralODEs. However, the proposed shooting augmentation introduces model approximation and involves approximating inference over auxiliary shooting variables. Hence the benefits of the shooting augmentation can be task specific, especially on short sequences. Comprehensive empirical studies across different types of tasks can be considered in future work.

\bibliography{references}

\begin{thebibliography}{55}
\providecommand{\natexlab}[1]{#1}
\providecommand{\url}[1]{\texttt{#1}}
\expandafter\ifx\csname urlstyle\endcsname\relax
  \providecommand{\doi}[1]{doi: #1}\else
  \providecommand{\doi}{doi: \begingroup \urlstyle{rm}\Url}\fi

\bibitem[{\"A}ij{\"o} and L{\"a}hdesm{\"a}ki(2009)]{aijo2009learning}
Tarmo {\"A}ij{\"o} and Harri L{\"a}hdesm{\"a}ki.
\newblock Learning gene regulatory networks from gene expression measurements
  using non-parametric molecular kinetics.
\newblock \emph{Bioinformatics}, 25\penalty0 (22):\penalty0 2937--2944, 2009.

\bibitem[Archambeau et~al.(2007)Archambeau, Cornford, Opper, and
  Shawe-Taylor]{archambeau2007gaussian}
Cedric Archambeau, Dan Cornford, Manfred Opper, and John Shawe-Taylor.
\newblock Gaussian process approximations of stochastic differential equations.
\newblock In \emph{Gaussian Processes in Practice}, pages 1--16. PMLR, 2007.

\bibitem[Bock and Plitt(1984)]{bock1984}
Hans Bock and Karl Plitt.
\newblock A multiple shooting algorithm for direct solution of optimal control
  problems.
\newblock In \emph{IFAC World Congress}, pages 242--247, 1984.

\bibitem[Bock(1983)]{bock1983recent}
Hans~Georg Bock.
\newblock Recent advances in parameter identification techniques for ode.
\newblock \emph{Numerical treatment of inverse problems in differential and
  integral equations}, pages 95--121, 1983.

\bibitem[Butcher and Goodwin(2008)]{butcher2008numerical}
John Butcher and Nicolette Goodwin.
\newblock \emph{Numerical methods for ordinary differential equations}.
\newblock Wiley Online Library, 2nd edition, 2008.

\bibitem[Calderhead et~al.(2008)Calderhead, Girolami, and
  Lawrence]{calderhead2008}
Ben Calderhead, Mark Girolami, and Neil Lawrence.
\newblock Accelerating {B}ayesian inference over nonlinear differential
  equations with {G}aussian processes.
\newblock In \emph{Advances in Neural Information Processing Systems}, 2008.

\bibitem[Chen et~al.(2018)Chen, Rubanova, Bettencourt, and
  Duvenaud]{chen2018neural}
Tian~Qi Chen, Yulia Rubanova, Jesse Bettencourt, and David Duvenaud.
\newblock Neural ordinary differential equations.
\newblock In \emph{Advances in Neural Information Processing Systems}, pages
  6571--6583, 2018.

\bibitem[Choromanski et~al.(2020)Choromanski, Davis, Likhosherstov, Song,
  Slotine, Varley, Lee, Weller, and Sindhwani]{choromanski2020}
Krzysztof Choromanski, Jared Davis, Valerii Likhosherstov, Xingyou Song,
  Jean-Jacques Slotine, Jacob Varley, Honglak Lee, Adrian Weller, and Vikas
  Sindhwani.
\newblock An {O}de to an {ODE}.
\newblock In \emph{Advances in Neural Information Processing Systems},
  volume~33, 2020.

\bibitem[Dandekar et~al.(2020)Dandekar, Chung, Dixit, Tarek, Garcia-Valadez,
  Vemula, and Rackauckas]{dandekar2020bayesian}
Raj Dandekar, Karen Chung, Vaibhav Dixit, Mohamed Tarek, Aslan Garcia-Valadez,
  Krishna~Vishal Vemula, and Chris Rackauckas.
\newblock Bayesian neural ordinary differential equations.
\newblock \emph{arXiv preprint arXiv:2012.07244}, 2020.

\bibitem[Diehl and Gros(2017)]{diehl2017}
Moritz Diehl and Sebastian Gros.
\newblock \emph{Numerical Optimal Control}.
\newblock University of Freiburg, 2017.

\bibitem[Doerr et~al.(2018)Doerr, Daniel, Schiegg, Duy, Schaal, Toussaint, and
  Sebastian]{doerr2018probabilistic}
Andreas Doerr, Christian Daniel, Martin Schiegg, Nguyen-Tuong Duy, Stefan
  Schaal, Marc Toussaint, and Trimpe Sebastian.
\newblock Probabilistic recurrent state-space models.
\newblock In \emph{International Conference on Machine Learning}, pages
  1280--1289, 2018.

\bibitem[Dondelinger et~al.(2013)Dondelinger, Husmeier, Rogers, and
  Filippone]{dondelinger2013ode}
Frank Dondelinger, Dirk Husmeier, Simon Rogers, and Maurizio Filippone.
\newblock {ODE} parameter inference using adaptive gradient matching with
  {G}aussian processes.
\newblock In \emph{Artificial Intelligence and Statistics}, pages 216--228,
  2013.

\bibitem[Duncker et~al.(2019)Duncker, Bohner, Boussard, and
  Sahani]{duncker2019learning}
Lea Duncker, Gergo Bohner, Julien Boussard, and Maneesh Sahani.
\newblock Learning interpretable continuous-time models of latent stochastic
  dynamical systems.
\newblock In \emph{International Conference on Machine Learning}, pages
  1726--1734. PMLR, 2019.

\bibitem[Eleftheriadis et~al.(2017)Eleftheriadis, Nicholson, Deisenroth, and
  Hensman]{eleftheriadis2017identification}
Stefanos Eleftheriadis, Tom Nicholson, Marc~Peter Deisenroth, and James
  Hensman.
\newblock Identification of {G}aussian process state space models.
\newblock In \emph{Advances in Neural Information Processing Systems}, pages
  5309--5319, 2017.

\bibitem[Ensinger et~al.(2021)Ensinger, Solowjow, Tiemann, and
  Trimpe]{ensinger2021symplectic}
Katharina Ensinger, Friedrich Solowjow, Michael Tiemann, and Sebastian Trimpe.
\newblock Symplectic gaussian process dynamics.
\newblock \emph{arXiv preprint arXiv:2102.01606}, 2021.

\bibitem[Frigola et~al.(2014)Frigola, Chen, and
  Rasmussen]{frigola2014variational}
Roger Frigola, Yutian Chen, and Carl Rasmussen.
\newblock Variational {G}aussian process state-space models.
\newblock In \emph{Advances in Neural Information Processing Systems}, 2014.

\bibitem[Girolami(2008)]{girolami2008bayesian}
Mark Girolami.
\newblock Bayesian inference for differential equations.
\newblock \emph{Theoretical Computer Science}, 408:\penalty0 4--16, 2008.

\bibitem[Grathwohl et~al.(2019)Grathwohl, Chen, Bettencourt, Sutskever, and
  Duvenaud]{grathwohl2019ffjord}
Will Grathwohl, Ricky T.~Q. Chen, Jesse Bettencourt, Ilya Sutskever, and David
  Duvenaud.
\newblock {FFJORD}: {F}ree-form continuous dynamics for scalable reversible
  generative models.
\newblock In \emph{ICLR}, 2019.

\bibitem[Haber and Ruthotto(2017)]{haber2017}
Eldad Haber and Lars Ruthotto.
\newblock Stable architectures for deep neural networks.
\newblock \emph{Inverse Problems}, 34:\penalty0 014004, 2017.

\bibitem[Heinonen and d'Alch{\'e}{-}Buc(2014)]{heinonen2014learning}
Markus Heinonen and Florence d'Alch{\'e}{-}Buc.
\newblock Learning nonparametric differential equations with operator-valued
  kernels and gradient matching.
\newblock Technical report, Universite d'Evry, 2014.

\bibitem[Heinonen et~al.(2018)Heinonen, Yildiz, Mannerstr{\"o}m, Intosalmi, and
  L{\"a}hdesm{\"a}ki]{heinonen2018learning}
Markus Heinonen, Cagatay Yildiz, Henrik Mannerstr{\"o}m, Jukka Intosalmi, and
  Harri L{\"a}hdesm{\"a}ki.
\newblock Learning unknown {ODE} models with {G}aussian processes.
\newblock In \emph{International Conference on Machine Learning}, pages
  1959--1968, 2018.

\bibitem[Henderson and Michailidis(2014)]{henderson2014network}
James Henderson and George Michailidis.
\newblock Network reconstruction using nonparametric additive {ODE} models.
\newblock \emph{PLOS ONE}, 9\penalty0 (4):\penalty0 e94003, 2014.

\bibitem[Hensman et~al.(2013)Hensman, Fusi, and Lawrence]{hensman2013gaussian}
James Hensman, Nicol{\`o} Fusi, and Neil Lawrence.
\newblock Gaussian processes for big data.
\newblock In \emph{Uncertainty in Artificial Intelligence}, pages 282--290,
  2013.

\bibitem[Hensman et~al.(2015)Hensman, Matthews, Filippone, and
  Ghahramani]{hensman2015mcmc}
James Hensman, Alexander~G Matthews, Maurizio Filippone, and Zoubin Ghahramani.
\newblock Mcmc for variationally sparse gaussian processes.
\newblock \emph{Advances in Neural Information Processing Systems},
  28:\penalty0 1648--1656, 2015.

\bibitem[Hirsch et~al.(2012)Hirsch, Smale, and Devaney]{hirsch2012differential}
Morris Hirsch, Stephen Smale, and Robert Devaney.
\newblock \emph{Differential equations, dynamical systems, and an introduction
  to chaos}.
\newblock Academic press, 2012.

\bibitem[Ialongo et~al.(2019)Ialongo, Van Der~Wilk, Hensman, and
  Rasmussen]{ialongo2019overcoming}
Alessandro~Davide Ialongo, Mark Van Der~Wilk, James Hensman, and Carl~Edward
  Rasmussen.
\newblock Overcoming mean-field approximations in recurrent gaussian process
  models.
\newblock In \emph{International Conference on Machine Learning}, pages
  2931--2940, 2019.

\bibitem[J{\o}rgensen et~al.(2020)J{\o}rgensen, Deisenroth, and
  Salimbeni]{jorgensen2020stochastic}
Martin J{\o}rgensen, Marc Deisenroth, and Hugh Salimbeni.
\newblock Stochastic differential equations with variational wishart
  diffusions.
\newblock In \emph{International Conference on Machine Learning}, pages
  4974--4983. PMLR, 2020.

\bibitem[Kingma and Ba(2014)]{kingma2014adam}
Diederik~P Kingma and Jimmy~Lei Ba.
\newblock Adam: A method for stochastic optimization.
\newblock In \emph{Proceedings of 3rd International Conference on Learning
  Representations}, 2014.

\bibitem[Kokotovic and Heller(1967)]{kokotovic1967direct}
Petar Kokotovic and James Heller.
\newblock Direct and adjoint sensitivity equations for parameter optimization.
\newblock \emph{IEEE Transactions on Automatic Control}, 12\penalty0
  (5):\penalty0 609--610, 1967.

\bibitem[Li et~al.(2020)Li, Wong, Chen, and Duvenaud]{li2020scalable}
Xuechen Li, Ting-Kam~Leonard Wong, Ricky~TQ Chen, and David Duvenaud.
\newblock Scalable gradients for stochastic differential equations.
\newblock In \emph{International Conference on Artificial Intelligence and
  Statistics}, pages 3870--3882. PMLR, 2020.

\bibitem[Lindelöf(1894)]{lindelof}
Ernst Lindelöf.
\newblock Sur l'application de la méthode des approximations successives aux
  équations différentielles ordinaires du premier ordre.
\newblock \emph{Comptes rendus hebdomadaires des séances de l'Académie des
  sciences}, 116:\penalty0 454--457, 1894.

\bibitem[Massaroli et~al.(2021)Massaroli, Poli, Sonoda, Suzuki, Park,
  Yamashita, and Asama]{massaroli2021differentiable}
Stefano Massaroli, Michael Poli, Sho Sonoda, Taiji Suzuki, Jinkyoo Park,
  Atsushi Yamashita, and Hajime Asama.
\newblock Differentiable multiple shooting layers.
\newblock \emph{Advances in Neural Information Processing Systems}, 34, 2021.

\bibitem[Mikheeva et~al.(2021)Mikheeva, Kazlauskaite, Hartshorne,
  Kjellstr{\"o}m, Ek, and Campbell]{mikheeva2021aligned}
Olga Mikheeva, Ieva Kazlauskaite, Adam Hartshorne, Hedvig Kjellstr{\"o}m,
  Carl~Henrik Ek, and Neill~DF Campbell.
\newblock Aligned multi-task gaussian process.
\newblock \emph{arXiv preprint arXiv:2110.15761}, 2021.

\bibitem[Osborne(1969)]{osborne1969}
Michael Osborne.
\newblock On shooting methods for boundary value problems.
\newblock \emph{Journal of Mathematical Analysis and Applications},
  27:\penalty0 417--433, 1969.

\bibitem[Pontryagin et~al.(1962)Pontryagin, Mishchenko, Boltyanskii, and
  Gamkrelidze]{pontryagin1962mathematical}
Lev Pontryagin, Evgenii Mishchenko, Vladimir Boltyanskii, and Revas
  Gamkrelidze.
\newblock \emph{The mathematical theory of optimal processes}.
\newblock Interscience Publishers, 1962.
\newblock Translation KN Trirogoff.

\bibitem[Rackauckas and Nie(2017)]{rackauckas2017differentialequations}
Christopher Rackauckas and Qing Nie.
\newblock Differentialequations.jl -- a performant and feature-rich ecosystem
  for solving differential equations in {J}ulia.
\newblock \emph{Journal of Open Research Software}, 5\penalty0 (1), 2017.

\bibitem[Rahimi and Recht(2007)]{rahimi2007random}
Ali Rahimi and Benjamin Recht.
\newblock Random features for large-scale kernel machines.
\newblock In \emph{Advances in Neural Information Processing Systems}, pages
  1177--1184, 2007.

\bibitem[Ramsay et~al.(2007)Ramsay, Hooker, Campbell, and
  Cao]{ramsay2007parameter}
Jim Ramsay, Giles Hooker, David Campbell, and Jiguo Cao.
\newblock Parameter estimation for differential equations: {A} generalized
  smoothing approach.
\newblock \emph{Journal of the Royal Statistical Society: Series B},
  69\penalty0 (5):\penalty0 741--796, 2007.

\bibitem[Rasmussen and Williams(2006)]{rasmussen2006gaussian}
Carl Rasmussen and Christopher Williams.
\newblock \emph{Gaussian processes for machine learning}.
\newblock The MIT Press, 2006.

\bibitem[Rubanova et~al.(2019)Rubanova, Chen, and Duvenaud]{rubanova2019}
Yulia Rubanova, Ricky T.~Q. Chen, and David Duvenaud.
\newblock Latent {ODE}s for irregularly-sampled time series.
\newblock In \emph{Advances in Neural Information Processing Systems}, 2019.

\bibitem[Stan(2021)]{stan}
Development~Team Stan.
\newblock Stan modeling language users guide and reference manual, mc-stan.org.
\newblock 2021.

\bibitem[Titsias(2009)]{titsias2009variational}
Michalis Titsias.
\newblock Variational learning of inducing variables in sparse {G}aussian
  processes.
\newblock In \emph{Artificial Intelligence and Statistics}, pages 567--574,
  2009.

\bibitem[Turan and J{\"a}schke(2021)]{turan2021multiple}
Evren~Mert Turan and Johannes J{\"a}schke.
\newblock Multiple shooting with neural differential equations.
\newblock \emph{arXiv preprint arXiv:2109.06786}, 2021.

\bibitem[Turner et~al.(2010)Turner, Deisenroth, and Rasmussen]{turner2010state}
Ryan Turner, Marc Deisenroth, and Carl Rasmussen.
\newblock State-space inference and learning with {G}aussian processes.
\newblock In \emph{Artificial Intelligence and Statistics}, pages 868--875,
  2010.

\bibitem[Tzen and Raginsky(2019)]{tzen2019neural}
Belinda Tzen and Maxim Raginsky.
\newblock Neural stochastic differential equations: Deep latent gaussian models
  in the diffusion limit.
\newblock \emph{arXiv preprint arXiv:1905.09883}, 2019.

\bibitem[Ustyuzhaninov et~al.(2020)Ustyuzhaninov, Kazlauskaite, Ek, and
  Campbell]{pmlr-v108-ustyuzhaninov20a}
Ivan Ustyuzhaninov, Ieva Kazlauskaite, Carl~Henrik Ek, and Neill Campbell.
\newblock Monotonic {G}aussian process flows.
\newblock In \emph{Artificial Intelligence and Statistics}, volume 108, pages
  3057--3067, 2020.

\bibitem[vanDomselaar and Hemker(1975)]{vandomselaar1975nonlinear}
B~vanDomselaar and Piet~W Hemker.
\newblock Nonlinear parameter estimation in initial value problems.
\newblock \emph{Stichting Mathematisch Centrum. Numerieke Wiskunde}, \penalty0
  (NW 18/75), 1975.

\bibitem[Varah(1982)]{varah1982spline}
James Varah.
\newblock A spline least squares method for numerical parameter estimation in
  differential equations.
\newblock \emph{SIAM Journal on Scientific and Statistical Computing},
  3\penalty0 (1):\penalty0 28--46, 1982.

\bibitem[Wang et~al.(2005)Wang, Hertzmann, and Fleet]{wang2005}
Jack Wang, Aaron Hertzmann, and David Fleet.
\newblock Gaussian process dynamical models.
\newblock In \emph{Advances in Neural Information Processing Systems}, 2005.

\bibitem[Wang et~al.(2008)Wang, Fleet, and Hertzmann]{4359316}
Jack Wang, David Fleet, and Aaron Hertzmann.
\newblock Gaussian process dynamical models for human motion.
\newblock \emph{IEEE Transactions on Pattern Analysis and Machine
  Intelligence}, 30\penalty0 (2):\penalty0 283--298, 2008.

\bibitem[Wenk et~al.(2020)Wenk, Abbati, Osborne, Sch{\"o}lkopf, Krause, and
  Bauer]{wenk2020odin}
Philippe Wenk, Gabriele Abbati, Michael Osborne, Bernhard Sch{\"o}lkopf,
  Andreas Krause, and Stefan Bauer.
\newblock {ODIN}: {ODE}-informed regression for parameter and state inference
  in time-continuous dynamical systems.
\newblock In \emph{AAAI Conference on Artificial Intelligence}, volume~34,
  pages 6364--6371, 2020.

\bibitem[Wilson et~al.(2020)Wilson, Borovitskiy, Terenin, Mostowsky, and
  Deisenroth]{wilson2020efficiently}
James Wilson, Viacheslav Borovitskiy, Alexander Terenin, Peter Mostowsky, and
  Marc Deisenroth.
\newblock Efficiently sampling functions from {G}aussian process posteriors.
\newblock In \emph{International Conference on Machine Learning}, pages
  10292--10302, 2020.

\bibitem[Yildiz et~al.(2019)Yildiz, Heinonen, and
  Lähdesmäki]{yildiz2019ode2vae}
Cagatay Yildiz, Markus Heinonen, and Harri Lähdesmäki.
\newblock {ODE2VAE}: Deep generative second order {ODE}s with {B}ayesian neural
  networks.
\newblock In \emph{Advances in Neural Information Processing Systems}, pages
  13412--13421, 2019.

\bibitem[Zhuang et~al.(2020)Zhuang, Dvornek, Li, Tatikonda, Papademetris, and
  Duncan]{zhuang2020}
Juntang Zhuang, Nicha Dvornek, Xiaoxiao Li, Sekhar Tatikonda, Xenophon
  Papademetris, and James Duncan.
\newblock Adaptive checkpoint adjoint method for gradient estimation in neural
  {ODE}.
\newblock In \emph{International Conference on Machine Learning}, pages
  11639--11649, 2020.

\bibitem[Zhuang et~al.(2021)Zhuang, Dvornek, Tatikonda, and
  Duncan]{zhuang2021mali}
Juntang Zhuang, Nicha Dvornek, Sekhar Tatikonda, and James Duncan.
\newblock {MALI}: {A} memory efficient and reverse accurate integrator for
  neural {ODE}s.
\newblock In \emph{ICLR}, 2021.

\end{thebibliography}

\newpage
\onecolumn
\thispagestyle{plain}

\hrule height 2pt
\begin{center}
\textbf{\large{Appendix : Variational multiple shooting for Bayesian ODEs with Gaussian processes}}
\end{center}
\hrule height 1pt

\makeatletter
\setcounter{equation}{0}
\setcounter{figure}{0}
\setcounter{table}{0}
\setcounter{page}{1}
\setcounter{section}{0}
\makeatletter
\renewcommand{\bibnumfmt}[1]{[S#1]}
\renewcommand{\citenumfont}[1]{S#1}
\renewcommand{\thesection}{S\arabic{section} }
\renewcommand*{\thesubsection}{S\arabic{section}.\arabic{subsection}}

\section{Detailed Derivations}
\subsection{Inference for the vanilla GPODE model}

\paragraph{The model.}
We consider the problem of inferring an ODE system
\begin{align}
    \y(t) &= \x(t) + \be \\ 
    \x(t) &= \x_0 + \int_0^t \f( \x(\tau)) d\tau \label{eq:supp_odeproblem}
\end{align}
from some noisy observations $\y(t)$ of the true system state $\x(t) \in \R^D$, whose evolution over time $t \in \R_+$ follows a differential equation vector field
\begin{align}
    \dot{\x}(t) &= \frac{d\x(t)}{dt} := \f(\x(t)), \qquad \f : \R^D \mapsto \R^D
\end{align}
starting from an initial state $\x_0 \in \R^D$. Our goal is to learn the underlying ODE vector field $\f$.

We propose a Gaussian process prior to the differential function
\begin{align}
    \f(\x) &\sim \GP(\0, k(\x, \x')).
\end{align}

Following \citet{titsias2009variational} for sparse inference of GPs using inducing variables, we augment the full model with inducing values $\U = (\bu_1, \ldots, \bu_M)^T \in R^{M \cross D}$ and inducing locations $\Z = (\z_1, \ldots, \z_M)^T  \in R^{M \cross D}$, which results in a low-rank GP
\begin{align}
    p(\U) &= \N(\U | \0, \K_{\Z \Z}) \\
    p(\f | \U) &= \N(\f | \A \mathrm{vec}(\U), \K_{\X \X} - \A \K_{\Z \Z} \A^T),
\end{align}
where $\X = (\x_1, \x_2, \ldots \x_{N'})^T \in \R^{N' \times D}$ collects all the intermediate state evaluations $\x(t_i)$ encountered along numerical approximation of the true continuous ODE integral \eqref{eq:supp_odeproblem}, $\f = (\f(\x_1)^T, \ldots, \f(\x_{N'})^T)^T \in \mathbb{R}^{N'D \times 1}$, $\K_{\X\X}$ is a block-partitioned matrix of size $N'D \times N'D$ with $D \times D$ blocks, so that block $(\K_{\X\X})_{i,j} = K(\x_i, \x_j)$, and $\A = \K_{\X\Z} \K^{-1}_{\Z\Z}$. 

\paragraph{The joint model} 
The joint probability of the model is
\begin{align}
    p(\Y,\f,\U, \x_0) &= \prod_{i=1}^{N} p(\y_i | \f, \x_0) p(\f|\U)p(\U) p(\x_0)\\
    &=  \prod_{i=1}^{N} \underbrace{p(\y_i | \f, \x_0)}_{\text{likelihood}} \underbrace{p(\f, \U)}_{\text{GP prior}} \underbrace{p(\x_0)}_{\text{initial state prior}},
\end{align}
where we assume a standard Gaussian prior $p(\x_0) = \N(\0, \eye)$ for the unknown initial state $\x_0$.

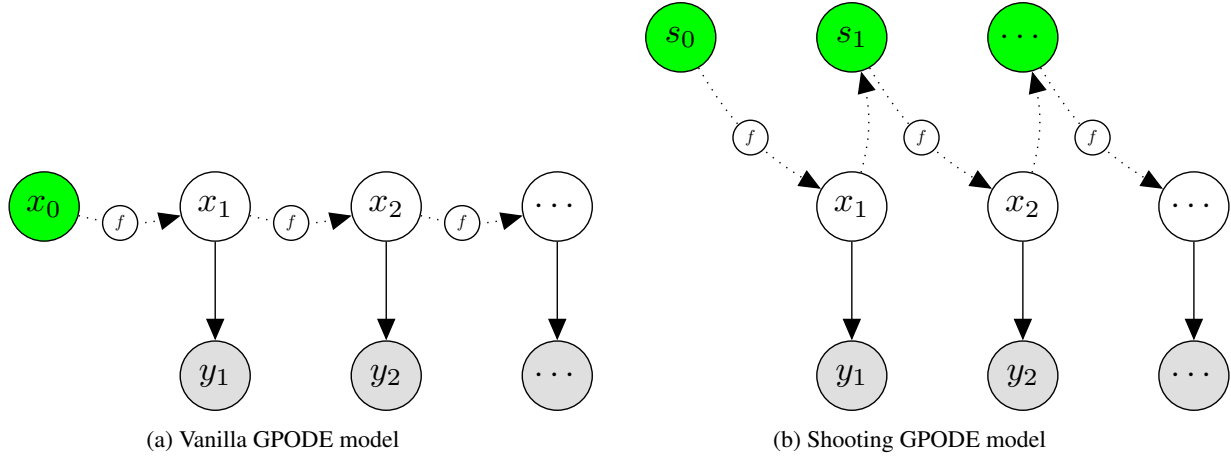
\begin{figure}[ht]
    \centering
    \begin{subfigure}[b]{0.42\columnwidth}
    \resizebox{1.1\columnwidth}{!}{
    \begin{tikzpicture}
     \node[latent][fill=green] (x0) {$x_0$}; 
     \node[latent,right=of x0] (x1) {$x_1$}; 
     \node[latent,right=of x1] (x2) {$x_2$}; 
     \node[latent,right=of x2] (xn) {$\cdots$}; 
     \node[obs,below=of x1] (y1) {$y_1$}; 
     \node[obs,right=of y1] (y2) {$y_2$}; 
     \node[obs,right=of y2] (yn) {$\cdots$}; 
     \path[->] (x0) edge[dotted,bend right=15] node[fill=white, anchor=center, pos=0.4, draw, circle, solid, scale=0.5] {$f$} (x1);
     \path[->] (x1) edge[dotted,bend right=15] node[fill=white, anchor=center, pos=0.4, draw, circle, solid, scale=0.5] {$f$} (x2);
     \path[->] (x2) edge[dotted,bend right=15] node[fill=white, anchor=center, pos=0.4, draw, circle, solid, scale=0.5] {$f$} (xn);
     \edge {x1} {y1};
     \edge {x2} {y2};
     \edge {xn} {yn};
    \end{tikzpicture}
    }
    \caption{Vanilla GPODE model}
    \end{subfigure}
    \hspace{10mm}
    \begin{subfigure}[b]{0.42\columnwidth}
    \resizebox{1.1\columnwidth}{!}{
    \begin{tikzpicture}
     \node[latent][fill=green]           (s0) {$s_0$}; 
     \node[latent,right=of s0][fill=green] (s1) {$s_1$}; 
     \node[latent,right=of s1][fill=green] (sn) {$\cdots$}; 
     \node[latent,below=of s1] (x1) {$x_1$}; 
     \node[latent,right=of x1] (x2) {$x_2$}; 
     \node[latent,right=of x2] (xn) {$\cdots$}; 
     \node[obs,below=of x1] (y1) {$y_1$}; 
     \node[obs,right=of y1] (y2) {$y_2$}; 
     \node[obs,right=of y2] (yn) {$\cdots$}; 
     \path[->] (s0) edge[dotted,bend right=15] node[fill=white, anchor=center, pos=0.5, draw, circle, solid, scale=0.5] {$f$} (x1);
     \path[->] (s1) edge[dotted,bend right=15] node[fill=white, anchor=center, pos=0.5, draw, circle, solid, scale=0.5] {$f$} (x2);
     \path[->] (sn) edge[dotted,bend right=15] node[fill=white, anchor=center, pos=0.5, draw, circle, solid, scale=0.5] {$f$} (xn);
     \path[->] (x1) edge[dotted,bend right=15](s1);
     \path[->] (x2) edge[dotted,bend right=15](sn);
     \edge {x1} {y1};
     \edge {x2} {y2};
     \edge {xn} {yn};
    \end{tikzpicture}
    }
    \caption{Shooting GPODE model}
    \end{subfigure}
    \hspace{10mm}
    \caption{Plate diagrams: latent random variables that are considered during model inference are shaded in green. The intermediate variables $\x_i$ (unshaded) are defined as deterministic transformations of the inferred variables (conditioned on the vectorfield). In the vanilla GPODE formulation (a), the initial state distribution $\x_0$ is integrated forward in time to match all the observations $\{\y_1, \y_2, \ldots, \y_N\}$ forming a full trajectory. The shooting version (b) splits the full trajectory into multiple subintervals. Every subinterval $i$ starts with an approximated state distribution $\s_i$, which is integrated forward to match the next observation $\y_{i+1}$. In addition, the state evolution from the previous shooting variable is matched to the variational shooting approximation at the current state.}
    \label{fig:plate}
\end{figure}

\paragraph{Inference.}
Our primary goal is to learn the vector field $\f$ by inferring the model posterior $p(\f,\U, \x_0 | \Y)$, which is intractable. We resort to stochastic variational inference \citet{hensman2013gaussian}, and introduce a factorized Gaussian posterior approximation  for the inducing variables across state dimensions 
\begin{align}
    q(\U) &= \prod_{d=1}^{D}\N(\bu_d|\m_d, \Q_d),
\end{align}
where, $\bu_d \in \R^M$ and  $\m_d \in \R^M, \Q_d \in \R^{M\times M}$ are the mean and covariance parameters of the variational Gaussian posterior approximation for the inducing variables. 
The Gaussian process posterior process with an inducing approximation can be written as
\begin{align}
    q(\f) &= \int p(\f|\U) q(\U) d\U \\
          &= \int \N\left(\f | \A \mathrm{vec}(\U), \K_{\X\X} - \A \K_{\Z\Z}\A^T \right) q(\U) d\U. \label{eq:supp_posterior_gp}
\end{align}
We also introduce posterior approximation for the initial state variable $\x_0$,
\begin{align}
    q(\x_0) = \N(\x_0 | \m_0, \S_0).
\end{align}
This results in a variational joint posterior approximation
\begin{align}
    q(\f,\U,\x_0) &= q(\f,\U)q(\x_0)\\ 
    &= p(\f|\U) q(\U) q(\x_0),
\end{align}

\paragraph{ELBO.}
With the above model specification, under variational inference of the posterior approximations, the evidence lower bound (ELBO) $\log p(\Y) \ge \L$ can be written as,
\begin{align}
    \L &= \iiint q(\f, \U, \x_0) \log \frac{p(\Y, \f, \U , \x_0)}{q(\f, \U, \x_0)} d\f d\U d\x_0 \\
    &= \iiint q(\f, \U, \x_0) \log \prod_{i=1}^{N} \underbrace{p(\y_i | \f, \x_0)}_{\L_y}  \frac{p(\f|\U)}{p(\f|\U)} \underbrace{\frac{p(\U)}{q(\U)}}_{\L_u}  \underbrace{\frac{p(\x_0)}{q(\x_0)}}_{\L_{\x_0}} d\f d\U d\x_0.
\end{align}
Hence the ELBO decomposes into three additive terms
\begin{align}
    \L &= \L_y + \L_u + \L_{\x_0},
\end{align}
where each term contains the (relevant parts of) expectation over $q(\f,\U,\x_0)$.

\paragraph{Likelihood term.}
The variational likelihood term $\L_y$ is an expectation of the likelihood wrt the variationally marginalized vectorfield posterior $q(\f)$, and the initial state distribution $q(\x_0)$,
\begin{align}
    \L_y &= \iint q(\f,\x_0) \log p(\y | \f, \x_0) d\f d\x_0 \\ 
         &= \sum_{i=1}^N \E_{q(\f, \x_0)} \log p(\y_i|\f, \x_0).
\end{align}
This term computes the likelihood $p(\y_i|\f, \x_0) = p(\y_i|\x_i)$ over ODE state solutions $\x_i = \x_0 + \int_0^{t_i} \f( \x(\tau)) d\tau$ for a single realization of the vector field $\f \sim p(\f)$ and the initial state $\x_0 \sim p(\x_0)$. Because of the non-linear integration $\x_0 \mapsto \x(t)$, we cannot solve this integral analytically. Instead, we resort to Monte Carlo integration by sampling ODE trajectories over different vector field realizations $\f \sim q(\f)$ and initial states $\x_0 \sim q(\x_0)$. In practice, this term can be approximated as  
\begin{align}
    \L_y &\approx \frac{1}{S}\sum_{s=1}^S \sum_{i=1}^N \log p(\y_i|\f^{(s)}, \x^{(s)}_0)
\end{align}
where we sum over $S$ reparameterized samples $\f^{(s)} \sim q(\f)$ and $\x^{(s)}_0 \sim q(\x_0)$.

\paragraph{Inducing KL.} This term corresponds to the KL divergence between variational posterior and the prior distribution of inducing values. This term can be derived analytically as the KL between multivariate Gaussians. 
\begin{align}
\L_u &= \int q(\U) \log \frac{p(\U)}{q(\U)} d\U \\
&= \sum_{d=1}^{D}\int q(\bu_d) \log \frac{p(\bu_d)}{q(\bu_d)} d\bu \\
&=- \sum_{d=1}^{D}\KL\left[q(\bu_d)||p(\bu_d)\right]
\end{align}

\paragraph{Initial state KL.} This term corresponds to the KL divergence between variational posterior and the prior distribution of the initial state. With an assumption of Gaussian prior and variational posterior, this term can also be derived analytically, 
\begin{align}
\L_{\x_0} &= \int q(\x_0) \log \frac{p(\x_0)}{q(\x_0)} d\x_0 \\
&= - \KL\left[q(\x_0)||p(\x_0)\right]
\end{align}

\paragraph{Complete ELBO.}
The full ELBO is then
\begin{align}
    \L &= \sum_{i=1}^N \E_{q(\f, \x_0)} \log p(\y_i | \f, \x_0) - \KL[ q(\U) \, || \, p(\U)] - \KL[ q(\x_0) \, || \, p(\x_0) ] \label{eq:supp_full_elbo}
\end{align}

\subsection{Decoupled sampling of GPODEs}
In this section, we provide details for simulating valid ODE trajectories from a GP vector field posterior of the form 
\begin{align}
    q(\bu) &= \N(\m, \Q), \\
    q(\f) &= \int p(\f|\bu) q(\bu) d\bu \\
          &= \int \N\left(\f | \A \bu, \K_{\X\X} - \A \K_{\Z\Z}\A^T \right) q(\bu) d\bu, \label{eq:supp_vf_posterior}
\end{align}
where $\A = \K_{\X\Z} \K^{-1}_{\Z\Z}$ and $\m \in \R^M, \Q \in \R^{M \times M}$ are the variational mean and covariance parameters of the Gaussian posterior approximation for inducing variables. For simplicity, we consider a scalar valued GP, but it is straightforward to extend this approach to vector-valued GPs.

A sparse GP posterior of the form \eqref{eq:supp_vf_posterior} can be decomposed into two parts using Matheron's rule (Corollary 2 \citet{ wilson2020efficiently}),
\begin{align}
\label{eq:supp_decoupled_conditional}
    \underbrace{f(\x)|\bu}_\texttt{posterior} &= \underbrace{f(\x)}_\texttt{prior} + \underbrace{k(\x,\Z)K(\Z,\Z)^{-1}(\bu - \f_{\Z}))}_\texttt{update}.
\end{align}

\citet{wilson2020efficiently} propose a decoupled sampling from the \texttt{posterior} by using different bases for the \texttt{prior} and \texttt{update} terms. In particular, they propose Fourier basis functions for the \texttt{prior} term and canonical basis for the \texttt{update} term respectively 
\begin{align}
\label{eq:supp_decoupled_bases}
    \underbrace{f(\x)|\bu}_\texttt{posterior} &\approx \underbrace{\sum_{i=1}^{F} w_i \phi_i(\x)}_\texttt{prior} + \underbrace{\sum_{j=1}^{M} \nu_j K(\x, \z_j)}_\texttt{update},
\end{align}
where we use $F$ Fourier bases $\phi_i(\cdot)$ with $w_i \sim \N(0,1)$ \citep{rahimi2007random} to represent the stationary prior, and function basis $K(\cdot,\z_j)$ for the posterior update with $\bnu = K(\Z,\Z)^{-1}(\bu - \bPhi \w)$, $\bPhi = \bphi(\Z) \in \R^{M \times F}, \w \in \R^{F}$. We can evaluate functions from the posterior \eqref{eq:supp_vf_posterior} in linear time at arbitrary locations.

For the experimental results presented in the paper, we use a squared exponential kernel for which we can compute the feature maps $\bphi_i(\x) = \sqrt{\frac{\sigma_f^2}{F}}(\cos \x^T\bo_i, \sin \x^T\bo_i)$ where $\bo_i$ is sampled proportional to the spectral density of the squared exponential kernel $\bo_i \sim \N(\0, \Lambda^{-1})$, $\Lambda$ is a diagonal matrix collecting lengthscale parameters of the kernel $\Lambda = \text{diag}(l_1^2,l_2^2,\ldots, l_D^2)$ and $\sigma_f^2$ is the signal variance parameter. In the case of the squared exponential kernel, this results in $2F$ feature maps $\bphi(\x) \in \R^{2F}$, for which we sample weights $\w \in \R^{2F}$ from the standard Normal  $w_i \sim \N(0,1)$. By fixing random samples of feature maps $\bphi(\cdot)$, corresponding weights $\w$ and inducing values $\bu$ for an ODE integration call, we can sample a unique ODE trajectory from a posterior vector field of the form \eqref{eq:supp_vf_posterior}.

\subsection{Probabilistic shooting formulation for GPODEs}
\paragraph{The model.}
We consider the problem of inferring an ODE system
\begin{align}
    \y(t) &= \x(t) + \be \\ 
    \x(t) &= \s_0 + \int_0^t \f( \x(\tau)) d\tau,
\end{align}
from some noisy observations $\y(t)$ of the true system state $\x(t) \in \R^D$, whose evolution over time $t \in \R_+$ follows a differential equation
\begin{align}
    \dot{\x}(t) &= \frac{d\x(t)}{dt} := \f(\x(t)), \qquad \f : \R^D \mapsto \R^D
\end{align}
starting from the initial state $\s_0 \in \R^D$. Our goal is to learn the underlying ODE vector field $\f$.

\paragraph{Shooting augmentation.}
We propose an augmented `shooting' ODE system
\begin{align}
    \y_i &= \x(t_i; \s_{i-1}) + \be \\
    \x(t_i; \s_{i-1}) &= \s_{i-1} + \int_{t_{i-1}}^{t_i} \f(\x(\tau)) d\tau  \label{eq:supp_shooting_ode} \\
    \s_i &= \x(t_i; \s_{i-1}) + \bxi \label{eq:supp_shooting_tolerence},
\end{align}
where we divide the state function $\x(t)$ into $N$ short segments, with the end state of $i^{\textit{th}}$ segment $\x(t_i; \s_{i-1})$ defining solutions to initial value problems \eqref{eq:supp_shooting_ode} starting from the corresponding shooting variables $\s_{i-1}$. These short shooting segments follow the same differential $\f$ as the original model. The augmented system is equivalent to the original ODE system, in the limit when the tolerance parameter $\bxi \to \0$.

We assume Gaussian distributions on both observation noise and tolerance parameters, resulting in the following distributions,
\begin{align}
    p(\y_i|\s_{i-1}) &=  \N(\y_i|\x(t_i; \s_{i-1}), \sigma^2_y \eye); \qquad \be \sim \N(\0, \sigma^2_y\eye), \label{eq:observation_prior}\\
    p(\s_i|\s_{i-1}) &= \N(\s_i|\x(t_i; \s_{i-1}), \sigma^2_\xi \eye); \qquad \bxi \sim \N(\0, \sigma^2_\xi\eye). \label{eq:supp_shooting_prior}
\end{align}

\paragraph{Gaussian process ODE.}
We propose a Gaussian process prior for the differential function
\begin{align}
    \f(\x) &\sim \GP(\0, k(\x, \x'))
\end{align}
In addition, we augment the full model with inducing values $\U = (\bu_1, \ldots, \bu_M)^T \in R^{M \cross D}$ and inducing locations $\Z = (\z_1, \ldots, \z_M)^T  \in R^{M \cross D}$, which results in a low-rank GP
\begin{align}
    p(\U) &= \N(\U | \0, \K_{\Z \Z}) \\
    p(\f | \U) &= \N(\f | \A \mathrm{vec}(\U), \K_{\X \X} - \A \K_{\Z \Z} \A^T),
\end{align}
where $\A = \K_{\X \Z} \K_{\Z \Z}^{-1}$.

\paragraph{The joint model.} 
The joint probability of the model is
\begin{align}
    p(\Y,\S,\f,\U) &= p(\Y|\S,\f)p(\S|\f)p(\f|\U)p(\U) \\
    &=  \prod_{i=1}^{N} \underbrace{p(\y_i | \s_{i-1}, \f)}_{\text{likelihood}} \prod\limits_{i=1}^{N-1}\underbrace{p(\s_{i} | \s_{i-1}, \f)}_{\text{shooting prior}} \underbrace{p(\s_0)}_{\text{initial state}} \underbrace{p(\f|\U)p(\U)}_{\text{GP prior}},
\end{align}
where $\S = (\s_0, \s_1, \ldots \s_{N-1})^T \in \R^{N \times D}$ collects all shooting variables. 

We also note that observations are at indices $1,\ldots,N$, while the shooting variables are always one behind the observations at $0, \ldots, N-1$ (see plate diagram \ref{fig:plate} (b)).

\paragraph{Inference.}
Our primary goal is to learn the vector field $\f$ by inferring the model posterior $p(\S,\f,\U | \Y)$, which is intractable. Similar to non-shooting GPODEs, we introduce a factorized Gaussian posterior approximation  for the inducing variables across state dimensions 
\begin{align}
    q(\U) &= \prod_{d=1}^{D}\N(\bu_d|\m_d, \Q_d),
\end{align}
where, $\bu_d \in \R^M$ and  $\m_d \in \R^M, \Q_d \in \R^{M\times M}$ are the mean and covariance parameters of the variational Gaussian posterior approximation for the inducing variables. 

The Gaussian process posterior process with an inducing approximation can be written as
\begin{align}
    q(\f) &= \int p(\f|\U) q(\U) d\U \\
          &= \int \N\left(\f | \A \mathrm{vec}(\U), \K_{\X\X} - \A \K_{\Z\Z}\A^T \right) q(\U) d\U.
\end{align}

Next, we introduce a factorized Gaussian posterior approximations for the shooting variables $\S$ as well,
\begin{align}
    q(\S) &= \prod\limits_{i=0}^{N-1} q(\s_i) = \prod\limits_{i=0}^{N-1} \N(\s_i | \a_i, \bS_i).
\end{align}
where, $\a_i \in \R^D$ and $\bS_i \in \R^{D\times D}$ are the mean and covariance parameters of the variational Gaussian posterior approximation for the shooting variables. 

This results in a variational joint posterior approximation
\begin{align}
    q(\S,\f,\U) &= q(\S) q(\f,\U)\\ 
    &= \prod\limits_{i=0}^{N-1} q(\s_i) p(\f|\U) q(\U).
\end{align}

\paragraph{ELBO.}
Under variational inference the posterior approximations $q$ are optimized to match the true posterior in the KL sense,
\begin{align}
    \argmin_q \: \KL\big[ q(\S,\f,\U) \, || \, p(\S,\f,\U | \Y) \big].
\end{align}
This is equivalent to maximizing the evidence lower bound (ELBO) $\log p(\Y) \ge \L$,
\begin{align}
    \L &= \iiint q(\S, \f, \U) \log \Bigg[ \frac{p(\Y, \S, \f, \U)}{q(\S, \f, \U)}\Bigg] d\S d\f d\U\\
    &= \iiint  q(\S, \f, \U) \log \Bigg[\prod_{i=1}^{N} p(\y_i | \s_{i-1}, \f) \cdot \prod_{i=1}^{N-1} \frac{p(\s_{i} | \s_{i-1}, \f)}{q(\s_{i})}  \cdot \frac{p(\s_0)}{q(\s_0)} \cdot \frac{p(\f,\U)}{q(\f,\U)} \Bigg]d\S d\f d\U \\
    &= \underbrace{\iint  q(\S)q(\f) \log \prod_{i=1}^{N} p(\y_i | \s_{i-1}, \f) d\S d\f}_{\L_y} + \underbrace{\iint q(\S)q(\f) \log \prod_{i=1}^{N-1} p(\s_{i} | \s_{i-1}, \f)d\S d\f}_{\L_{sc}} \nonumber \\
    &\qquad  \underbrace{-  \int q(\S)  \log \prod_{i=1}^{N-1} q(\s_{i}) d\S }_{\L_{se}}+ \underbrace{\int  q(\s_0) \log \frac{p(\s_0)}{q(\s_0)} d\s_0}_{\L_0}  + \underbrace{\int  q(\U) \log \frac{p(\U)}{q(\U)} d\U}_{\L_u}   \\    
\end{align}
which results in the ELBO decomposing into four additive terms
\begin{align}
    \L &= \L_y + \L_{sc} + \L_{se} + \L_0 + \L_u,
\end{align}
where each term contains the (relevant parts of) expectation over $q(\S,\f,\U)$.

\paragraph{Likelihood term.}
The variational likelihood term $\L_y$ is an expectation of the likelihood under the posteriors of shooting variables $q(\S)$ and the posterior vectorfield $q(\f)$,
\begin{align}
    \L_y &= \iint  q(\S)q(\f) \log \prod_{i=1}^{N} p(\y_i | \s_{i-1}, \f) d\S d\f\\ 
         &= \sum_{i=1}^N \iint q(\s_{i-1}) q(\f) \log p(\y_i | \s_{i-1}, \f) d\s_{i-1} d\f \\ 
         &= \sum_{i=1}^N \E_{q(\s_{i-1})q(\f)} \Big[\log p(\y_i|\s_{i-1}, \f) \Big].
\end{align}
We can evaluate this term with Monte Carlo integration by taking reparameterized samples from the posteriors $\f^{(s)} \sim q(\f)$ and $\s^{(s)}_{i-1} \sim q(\s_{i-1})$ as below 
\begin{align}
    \L_y &= \sum_{i=1}^N \E_{q(\s_{i-1},\f)} \Big[\log p(\y_i|\s_{i-1}, \f) \Big]\\
    \L_y &\approx \frac{1}{S}\sum_{s=1}^S \sum_{i=1}^N \Big[\log p(\y_i|\x^{(s)}_i)\Big], 
\end{align}
where $\x^{(s)}_i$ is defined as solution to the following initial value problem,
\begin{align}
    \x^{(s)}_i := \x^{(s)}(t_i; \s_{i-1}) &= \s^{(s)}_{i-1} + \int_{t_{i-1}}^{t_i} \f^{(s)}(\x(\tau)) d\tau.
\end{align}

\paragraph{Shooting cross-entropy term.}
This term computes the cross-entropy between the prior specification for the shooting variables under the ODE evolution $p(\s_{i} | \s_{i-1}, \f)$, and the point-wise approximations $q(\s_i)$,
\begin{align}
    \L_{se} &= \iint q(\S)q(\f) \Big[ \log \prod_{i=1}^{N-1} p(\s_{i} | \s_{i-1}, \f)\Big] d\S d\f \\ 
    &= \iint q(\s_{N-1}) \cdots q(\s_1) q(\s_0) q(\f) \Big[ \log p(\s_{N-1}|\s_{N-2}, \f) \cdots p(\s_1|\s_0, \f)\Big] d\S d\f\\ 
    &= \sum_{i=1}^{N-1} \iint q(\f) q(\s_i) q(\s_{i-1}) \Big[ \log p(\s_i | \s_{i-1}, \f)\Big] d\s_{i-1} d\s_i d\f \\  
    &= \sum_{i=1}^{N-1} \E_{q(\s_{i}, \s_{i-1},\f)} \Big[ \log p\left(\s_i | \s_{i-1}, \f \right) \Big] .
\end{align}

This term can also be numerically estimated with Monte Carlo integration using posterior samples $\f^{(s)} \sim q(\f)$, $\s^{(s)}_{i-1} \sim q(\s_{i-1})$ and  $\s^{(s)}_{i} \sim q(\s_{i})$
\begin{align}
    \L_{se} &= \sum_{i=1}^{N-1} \E_{q(\s_{i}, \s_{i-1},\f)} \Big[ \log p\left(\s_i | \s_{i-1}, \f \right) \Big] \\
    &\approx \frac{1}{S}\sum_{s=1}^S \sum_{i=1}^{N-1} \log p\left(\s^{(s)}_i \big| \x^{(s)}_i \right), \\
     \x^{(s)}_i := \x^{(s)}(t_i; \s_{i-1}) &= \s^{(s)}_{i-1} + \int_{t_{i-1}}^{t_i} \f^{(s)}(\x(\tau)) d\tau.
\end{align}

\paragraph{Shooting entropy term.}
This term computes the entropy of the posterior approximations for shooting variables $q(\s_{i})$. Since we assume factorized Gaussian approximations, this term can be simplified analytically as the sum of Gaussian entropy. 
\begin{align}
    \L_{se} &= - \int q(\S)  \log \prod_{i=1}^{N-1} q(\s_{i}) d\S \\ 
    &= - \sum_{i=1}^{N-1} \E_{q(\s_i)}  \Big[ \log q(\s_{i})  \Big].
\end{align}

\paragraph{Initial state KL term.} This term corresponds to the KL divergence between variational posterior and the prior distribution of the initial state. With the assumption of Gaussian prior and variational posterior, this term can also be derived analytically, 
\begin{align}
\L_0 &= \int q(\s_0) \log \frac{p(\s_0)}{q(\s_0)} d\s_0 \\
&= - \KL\left[q(\s_0)||p(\s_0)\right].
\end{align}

\paragraph{Inducing KL term.} This term corresponds to the KL divergence between variational posterior and prior distribution of inducing values. This term can also be derived analytically as the KL between multivariate Gaussians. 
\begin{align}
\L_u &= \int q(\U) \log \frac{p(\U)}{q(\U)} d\U \\
&= \sum_{d=1}^{D}\int q(\bu_d) \log \frac{p(\bu_d)}{q(\bu_d)} d\bu \\
&=- \sum_{d=1}^{D}\KL\left[q(\bu_d)||p(\bu_d)\right].
\end{align}

\paragraph{Complete ELBO.}
The full ELBO is then
\begin{align}
    \L &= \L_y + \L_{sc} + \L_{se} + \L_0 + \L_u \\
&= \sum_{i=1}^N \E_{q(\s_{i-1},\f)} \Big[\log p(\y_i|\s_{i-1}, \f) \Big] + \sum_{i=1}^{N-1} \E_{q(\s_{i}, \s_{i-1},\f)} \Big[ \log p\left(\s_i | \s_{i-1}, \f \right) \Big]\nonumber\\
&\quad  - \sum_{i=1}^{N-1} \E_{q(\s_i)}  \Big[ \log q(\s_{i})  \Big] - \KL[q(\s_0) \, || \, p(\s_0)] - \KL[ q(\U) \, || \, p(\U)]
\end{align}
which in practice is numerically estimated with Monte Carlo integration
\begin{align}
\L &\approx \frac{1}{S}\sum_{s=1}^S \sum_{i=1}^N \Big[\log p(\y_i|\x^{(s)}_i)\Big] + \frac{1}{S}\sum_{s=1}^S \sum_{i=1}^{N-1} \log p\left(\s^{(s)}_i \big| \x^{(s)}_i \right) \nonumber\\
 &\quad - \sum_{i=1}^{N-1} \E_{q(\s_i)}  \Big[ \log q(\s_{i})  \Big] - \KL[q(\s_0) \, || \, p(\s_0)] - \KL[ q(\U) \, || \, p(\U)]
\end{align}
where $\f^{(s)} \sim q(\f)$, $\s^{(s)}_{i-1} \sim q(\s_{i-1})$, $\s^{(s)}_{i} \sim q(\s_{i})$ and 
\begin{align}
     \x^{(s)}_i &:= \x^{(s)}(t_i; \s_{i-1}) = \s^{(s)}_{i-1} + \int_{t_{i-1}}^{t_i} \f^{(s)}(\x(\tau)) d\tau.
\end{align}

\section{Experimental Details}
\begin{algorithm}[th]
\caption{GPODEs : Bayesian inference of ODEs using Gaussian processes}
\label{alg:gpode}
\begin{algorithmic}
   \STATE {\bfseries Inputs:} 
   \STATE {} \hspace{5mm}  - Observed states $\mathbf{Y}$, observation time sequence $\mathbf{t}$.
   \STATE {\bfseries Initialize hyperparameters:} 
   \STATE {} \hspace{5mm}  - Kernel parameters $\mathbf{\theta}$, likelihood parameters, inducing locations $\mathbf{Z}$.
   \STATE {\bfseries Initialize variational parameters:} 
   \STATE {} \hspace{5mm}  - Parameters of $q(\mathbf{U}) =  \mathcal{N}(\mathbf{m}, \mathbf{Q})$.
   \STATE {} \hspace{5mm}  - Parameters of $q(\mathbf{x_0}) = \mathcal{N}(\mathbf{a_0}, \mathbf{\Sigma_0})$.
   \STATE {\bfseries Optimization:} 
   \FOR {every optimization step}
    \STATE (1) Sample a function $\mathbf{f}$ from the ODE posterior in \eqref{eq:supp_posterior_gp} by taking following samples:
    \STATE {} \hspace{5mm}  - Parameters of Fourier bases $\mathbf{\omega_\theta}$  proportional to the spectral density of GP kernel,
    \STATE {} \hspace{5mm}  - Weights $\mathbf{w} \sim \mathcal{N}(\mathbf{0}, \textbf{\textrm{I}})$,
    \STATE {} \hspace{5mm}  -  Sample from the inducing posterior $\mathbf{U} \sim \mathcal{N}(\mathbf{m}, \mathbf{Q})$.
    \STATE (2) Sample initial state $\mathbf{x}_0 \sim \mathcal{N}(\mathbf{a_0}, \mathbf{\Sigma_0})$.
    \STATE (3) Compute predicted states $\hat{\mathbf{Y}} = \textrm{ODEsolve}(\mathbf{f}, \mathbf{x_0}, \mathbf{t})$.
    \STATE (4) Compute ELBO from \eqref{eq:supp_full_elbo} : $\textrm{likelihood}(\mathbf{Y}, \hat{\mathbf{Y}})$, 
    $\textrm{KL}[q(\mathbf{U})||p(\mathbf{U})]$, $\textrm{KL}[q(\mathbf{x_0})||p(\mathbf{x_0})]$.
   \STATE (5) Update all parameters with stochastic gradients of ELBO.
   \ENDFOR
\end{algorithmic}
\end{algorithm}

\subsection{Optimization setup}
We use Adam \citep{kingma2014adam} optimizer and jointly train all the variational parameters and hyperparameters.
The complete list of optimized parameters, along with additional method-specific details, are given below.

\paragraph{Vanilla GPODE model.} We use `whitened' representation for the inducing variables and optimize following parameters against the evidence lowerbound (see algorithm \ref{alg:gpode}). 
\begin{itemize}[noitemsep,nolistsep]
    \item Variational parameters:
    \begin{itemize}
        \item Inducing variables $q(\U)$,  initial states $q(\x_0)$
    \end{itemize}
    \item Hyperparameters:
    \begin{itemize}
        \item Inducing locations $\Z$
        \item Likelihood parameters: scale parameter for the Gaussian likelihood
        \item Kernel parameters: length scales and signal variance parameters in case of squared exponential kernel
    \end{itemize}
\end{itemize}

\paragraph{Shooting GPODE model.} We use `whitened' representation for inducing variables and optimize the following parameters against the evidence lower bound. 
\begin{itemize}[noitemsep,nolistsep]
    \item Variational parameters:
    \begin{itemize}
        \item Inducing variables $q(\U)$, shooting states $q(\S)$
    \end{itemize}
    \item Hyperparameters:
    \begin{itemize}
        \item Inducing locations $\Z$
        \item Likelihood parameters: scale parameter for the Gaussian likelihood
        \item Kernel parameters: length scales and signal variance parameters in case of squared exponential kernel
    \end{itemize}
\end{itemize}

\paragraph{npODE model.} We use `whitened' representation for inducing variables, maximum a posteriori (MAP) objective, and optimize following parameters:
\begin{itemize}[noitemsep,nolistsep]
    \item Inducing values $\U$ and locations $\Z$.
    \item Likelihood parameters: scale parameter for the Gaussian likelihood.
    \item Kernel parameters: length scales and signal variance parameters in case of the squared exponential kernel. 
\end{itemize}

\paragraph{NeuralODE model.} We use \texttt{tanh} activation and a fully connected block with one hidden layer having $32$ units in Van der Pol/ Fitz-Hugh Nagumo experiments. In MoCap experiments, we try one/two hidden layers with $64$/$128$ hidden units, and report the best results. All the network parameters were optimized against \texttt{MSE} loss.

\paragraph{Bayesian NeuralODE model.} We utilized the codebase \footnote{\url{https://github.com/RajDandekar/MSML21_BayesianNODE}} provided by \cite{dandekar2020bayesian} for training Bayesian version of NeuralODEs. We used networks with one hidden layer and $32$ units VDP/FHN experiments and performed posterior sampling with HMC. In case of experiments with long sequences (shooting illustration on VDP and MocCap) the HMC sampling had convergence issues, hence we performed variational inference instead. In case of MoCap experiments, we tried networks with two hidden layers and $64$/$128$ hidden units, and performed mean-field variational inference.

\subsection{Additional details on the inducing variables}
\paragraph{‘Whitening’ the inducing variables.} While performing sparse inference for GPs using inducing variables, it is a common practice to use noncental parameterization $\tilde{\mathbf{U}} = \mathbf{L_\theta U}$ where $\mathbf{L_\theta} \mathbf{L_\theta}^T =\mathbf{K_\theta}(\mathbf{Z},\mathbf{Z})$ \citep{hensman2015mcmc}. Such a reparametrization turns the inference for $\mathbf{U}$ with prior $\mathcal{N}(\mathbf{0}, \mathbf{K_{ZZ}})$ into inference for $\tilde{\mathbf{U}}$ with isotropic Gaussian prior $\mathcal{N}(\mathbf{0}, \textbf{\textrm{I}})$. This generally improves the optimization performance by decorrelating the latent parameters from each other.

\paragraph{Initializing inducing variables using data gradients.} In case of sparse Gaussian process model with inducing variables, we initialize the vector field with empirical gradients from the observed data. We first initialize inducing locations $\Z$ as \texttt{kmeans} cluster centers of observations $\Y$. Next we compute empirical gradient estimates, $\dot{\Y} = (\y_2-\y_1,\y_3-\y_2, \ldots, \y_{N} - \y_{N-1})$ at locations $\tilde{\Y} = (\y_1,\y_2, \ldots, \y_{N-1})$ and initialize inducing values $\U$ as the GP mean interpolation of empirical gradients at inducing locations. 
\begin{align}
    \U &=  \Delta t \cdot K(\Z, \tilde{\Y})K(\tilde{\Y}, \tilde{\Y})^{-1} \dot{\Y},
\end{align}
where $\Delta t$ is the time difference between two consecutive observations in the dataset.  

\subsection{Additional details on the CMU MoCap experiment}
\paragraph{Details on the dataset.} The dataset used in this experiment was obtained from \url{http://mocap.cs.cmu.edu/}. The database consists of sensor recordings of multiple activities for different subjects in \texttt{.amc} files. We selected three subjects with the most number of walking or running sequences: subjects \texttt{09}, \texttt{35}, and \texttt{39}. The \texttt{.amc} files considered for train, validation and test purposes are given in table \ref{table:supp_mocap_splits}. The training sequences and their lengths were selected to include at least one full cycle of the dynamics while learning the model. The observation sequence lengths for training/test/validation splits are reported in table \ref{table:supp_mocap_data}.

\paragraph{Details on the PCA} In the CMU MoCap experiment, we project the data from $D$ dimensional observation-space to $K<D$ dimensional latent-space using eigenvectors corresponding to top-$K$ eigenvalues. The ODE model is then learnt in the latent-space and model predictions are projected back into the observation-space using $K$ eigenvectors. We refer to this as `inverting the PCA' in the main text.

\begin{table}
\caption{For each subject (a), we report the activity considered for the experiment (b), the data split train/validation/test (c), the number of sequences considered for the corresponding split (d), and the files used in the corresponding split (e).}
\centering
\resizebox{0.8\columnwidth}{!}{
\begin{tabular}{l c c c c}
\toprule
 (a) subject & (b) activity & (c) split & (d) \# sequences & (e) files\\
\midrule
\multirow{ 2}{*}{subject \texttt{09}} & \multirow{ 2}{*}{running} &
train & 6 & \shortstack{\texttt{05.amc}, \texttt{06.amc}, \texttt{07.amc}, \\ \texttt{08.amc}, \texttt{09.amc}, \texttt{11.amc}}\\
 \cmidrule(lr){3-5} 
& & validation & 2 & \texttt{01.amc}, \texttt{02.amc}\\
\cmidrule(lr){3-5} 
& & test & 2 & \texttt{03.amc}, \texttt{04.amc}\\
\cmidrule(lr){1-5} 
\multirow{ 2}{*}{subject \texttt{35}} & \multirow{ 2}{*}{walking} &
train & 16 & \shortstack{\texttt{01.amc}, \texttt{02.amc}, \texttt{03.amc}, \texttt{04.amc}, \\
\texttt{05.amc}, \texttt{06.amc}, \texttt{07.amc}, \texttt{08.amc}, \\
\texttt{09.amc}, \texttt{10.amc}, \texttt{11.amc}, \texttt{12.amc}, \\
\texttt{13.amc}, \texttt{14.amc}, \texttt{15.amc}, \texttt{16.amc} }\\
\cmidrule(lr){3-5} 
& & validation & 3 & \texttt{28.amc}, \texttt{29.amc}, \texttt{30.amc} \\
\cmidrule(lr){3-5} 
& & test & 4 & \texttt{31.amc}, \texttt{32.amc}, \texttt{33.amc}, \texttt{34.amc}\\
\cmidrule(lr){1-5} 
\multirow{ 2}{*}{subject \texttt{39}} & \multirow{ 2}{*}{walking} &
train & 6 & \shortstack{\texttt{01.amc}, \texttt{02.amc}, \texttt{07.amc}, \\
\texttt{08.amc}, \texttt{09.amc}, \texttt{10.amc}}\\
\cmidrule(lr){3-5} 
& & validation & 2 & \texttt{03.amc}, \texttt{04.amc} \\
\cmidrule(lr){3-5} 
& & test & 2 & \texttt{05.amc}, \texttt{06.amc}\\
\bottomrule
\end{tabular}
}
\label{table:supp_mocap_splits}
\end{table}

\begin{table}[H]
\caption{For each subject (a), we report the experiment type (b), the data split train/validation/test (c), and the number of observations considered for the corresponding split.}
\centering
\resizebox{0.5\columnwidth}{!}{
\begin{tabular}{l c c c}
\toprule
 (a) subject & (b) experiment & (c) split & (d) sequence length\\
\midrule
\multirow{ 6}{*}{subject \texttt{09}} & \multirow{ 3}{*}{short} &
train & 50\\
& & validation & 120\\
& & test & 120\\
\cmidrule(lr){2-4}
& \multirow{ 3}{*}{long} &
train & 100\\
& & validation & 120\\
& & test & 120\\
\cmidrule(lr){1-4}
\multirow{ 6}{*}{subject \texttt{35}} & \multirow{ 3}{*}{short} &
train & 50\\
& & validation &300\\
& & test & 300\\
\cmidrule(lr){2-4}
& \multirow{ 3}{*}{long} &
train & 250\\
& & validation & 300\\
& & test & 300\\
\cmidrule(lr){1-4}
\multirow{ 6}{*}{subject \texttt{39}} & \multirow{ 3}{*}{short} &
train & 100\\
& & validation &300\\
& & test & 300\\
\cmidrule(lr){2-4}
& \multirow{ 3}{*}{long} &
train & 250\\
& & validation & 300\\
& & test & 300\\
\bottomrule
\end{tabular}
}
\label{table:supp_mocap_data}
\end{table}

\begin{figure*}[t]
    \centering
    \includegraphics[width=0.8\textwidth]{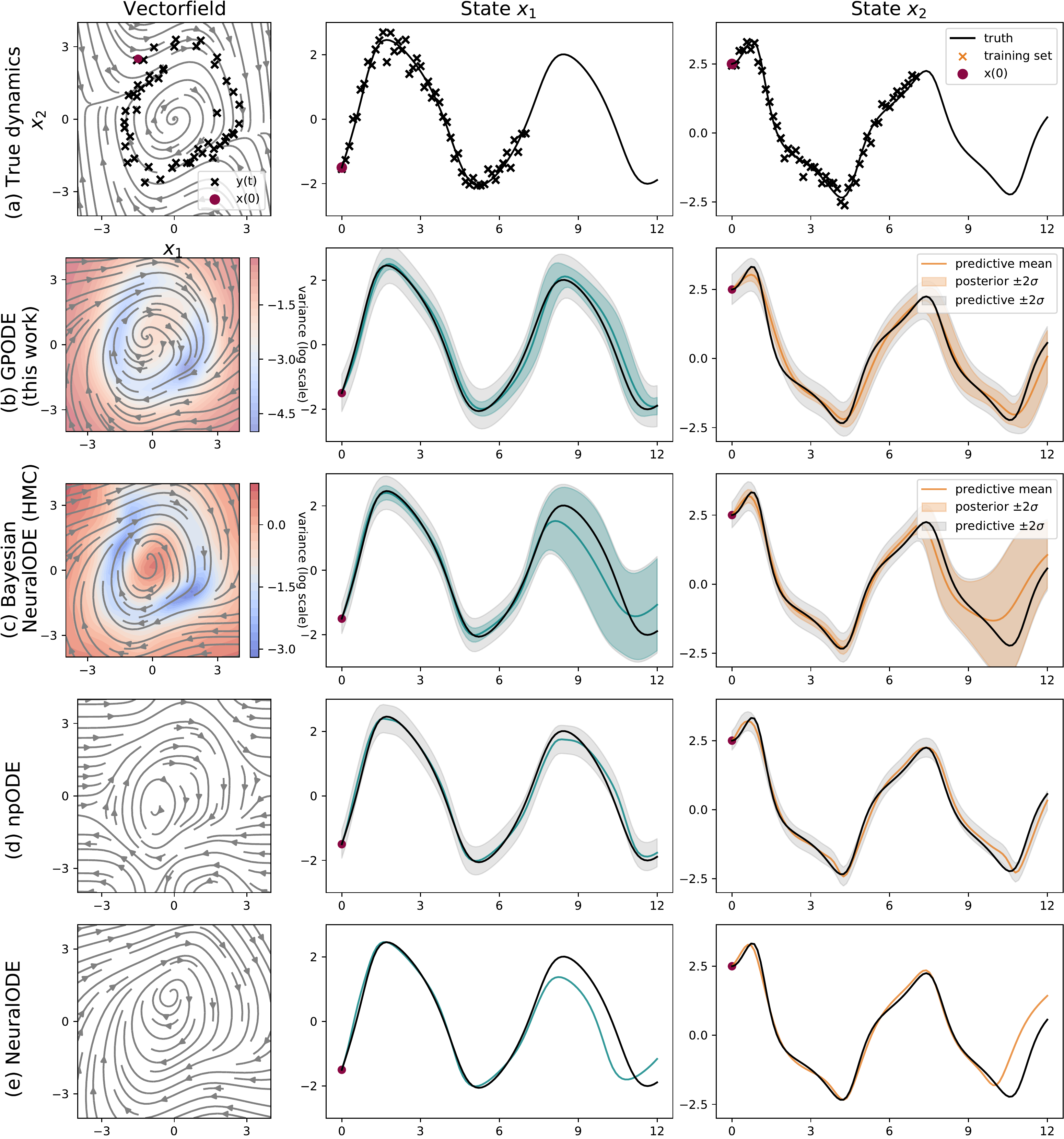}
    \caption{Learning the 2D Van der Pol dynamics on irregularly sampled observations \textbf{(a)} with alternative methods \textbf{(b-d)}. Column 1 shows the vector fields while columns 2 and 3 show the state trajectories $x_1(t)$ and $x_2(t)$. GPODE learns the posterior accurately.}
    \label{fig:vdp_illustration_irregular}
\end{figure*}

\begin{figure*}[t]
    \centering
    \begin{subfigure}[b]{0.45\columnwidth}
    \centering
    \includegraphics[width=1.0\textwidth]{plots/convergence_runtime.pdf}
    \caption{Convergence across wall-clock time}
    \end{subfigure}
    \qquad
    \begin{subfigure}[b]{0.45\columnwidth}
    \centering
    \includegraphics[width=1.0\textwidth]{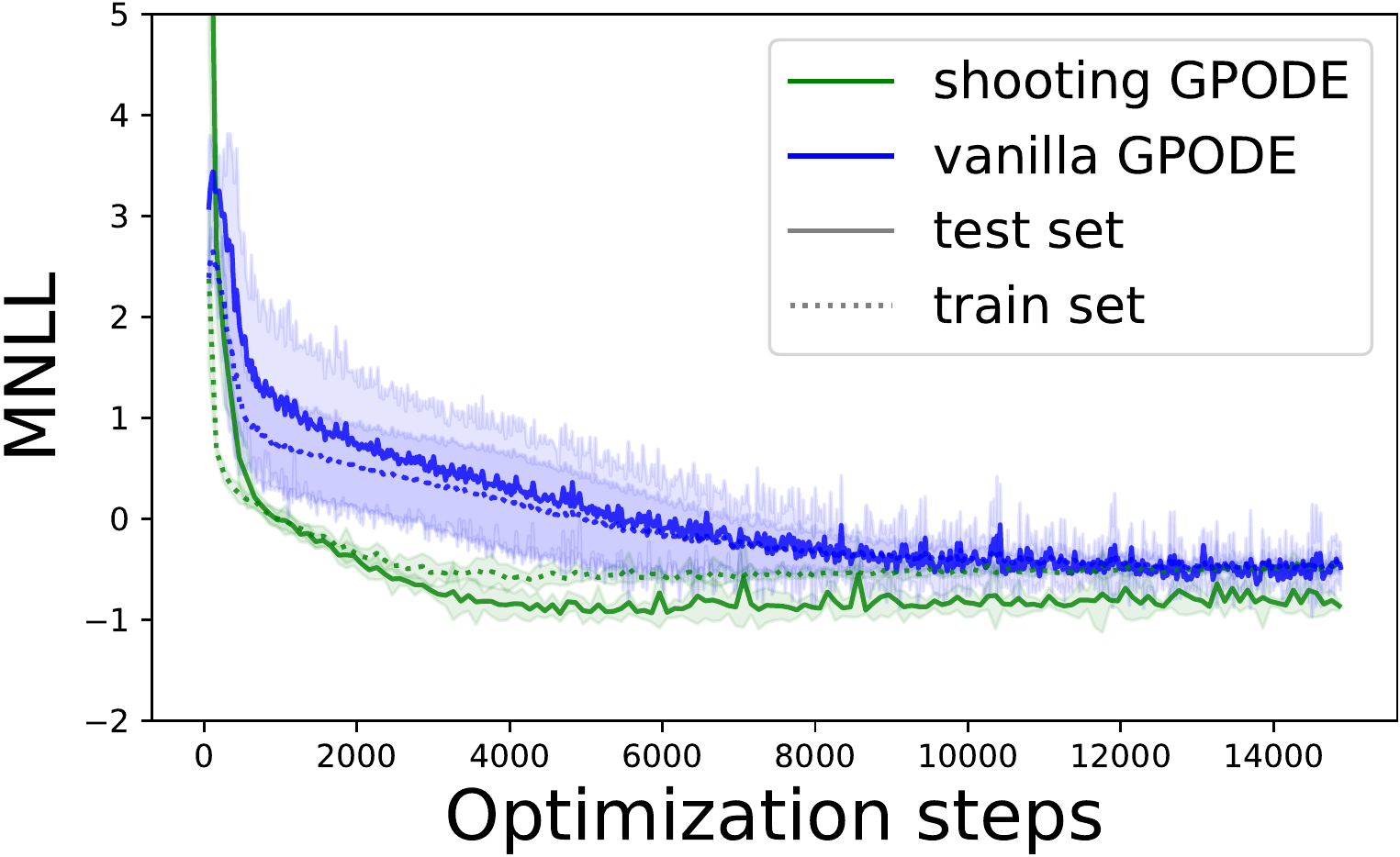}
    \caption{Convergence across gradient steps during optimization}
    \end{subfigure}
    \caption{Optimization efficiency with GPODE models.}
\end{figure*}

\end{document}